\pdfoutput=1

\documentclass[11pt]{article}

\usepackage{acl}

\usepackage{times}
\usepackage{latexsym}
\usepackage{footnote}
\usepackage{graphicx}  
\usepackage{subfig}  
\usepackage{booktabs}  
\usepackage{multirow}  
\usepackage{arydshln}  
\usepackage{amsmath,amssymb,amsfonts,amsthm,bm,bbm}  

\usepackage[T1]{fontenc}

\usepackage[utf8]{inputenc}

\usepackage{microtype}

\makeatletter
\newcommand\footnoteref[1]{\protected@xdef\@thefnmark{\ref{#1}}\@footnotemark}
\makeatother
\title{Graph Neural Network Policies and Imitation Learning \\
for Multi-Domain Task-Oriented Dialogues}

\author{Thibault Cordier\textsuperscript{*,1,2}, Tanguy Urvoy\textsuperscript{2}, Fabrice Lefèvre\textsuperscript{1}, Lina M. Rojas-Barahona\textsuperscript{2} \\
  \textsuperscript{1}LIA - Avignon University, Avignon, France \\
  \textsuperscript{2}Orange Labs, Lannion, France \\
  \small\texttt{thibault.cordier@alumni.univ-avignon.fr} \\
  \small\texttt{fabrice.lefevre@univ-avignon.fr} \\
  \small\texttt{\{thibault.cordier, linamaria.rojasbarahona, tanguy.urvoy\}@orange.com} \\
}

\begin{document}
\maketitle
\begin{abstract}
Task-oriented dialogue systems are designed to achieve specific goals while conversing with humans. In practice, they may have to handle simultaneously several domains and tasks. The dialogue manager must therefore be able to take into account domain changes and plan over different domains/tasks in order to deal with multi-domain dialogues. However, learning with reinforcement  in such context becomes difficult because the state-action dimension is larger while the reward signal remains scarce. 
Our experimental results suggest that structured policies based on graph neural networks combined with different degrees of imitation learning can effectively handle multi-domain dialogues. The reported experiments underline the benefit of structured policies over standard policies. 

\end{abstract}


\section*{Introduction}

\begin{figure*}[!ht]
  \begin{center}
    \subfloat[Domain-selection module.]{
        \includegraphics[width=0.37\textwidth]{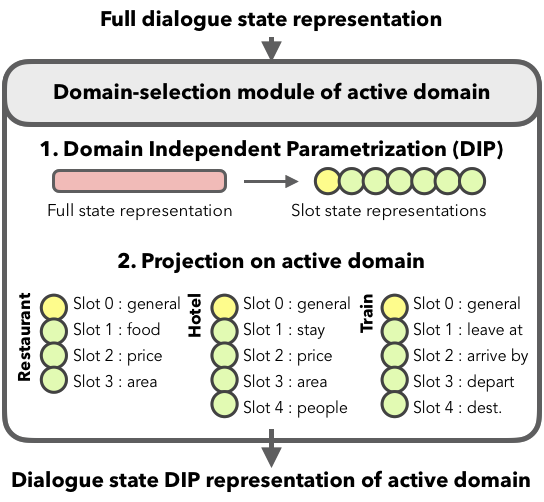}
        \label{sub:module_selection}
    }
    \hspace{0.05\textwidth}
    \subfloat[Domain-specific decision module.]{
        \includegraphics[width=0.37\textwidth]{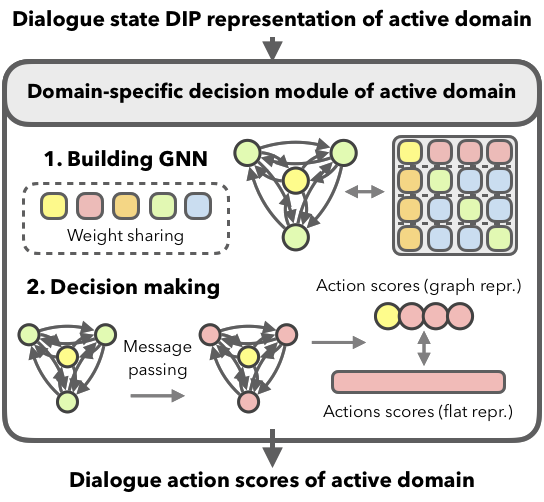}
        \label{sub:module_decision}
    }
  \end{center}
  \caption{GNN policy for multi-domain dialogues with hierarchical decision making and weight sharing.}
  \label{fig:proposal}
\end{figure*}


Task-oriented dialogue systems are designed to achieve specific goals while conversing with humans. They can help with various tasks in different domains, such as seeking and booking a restaurant or a hotel~\citep{zhu2020convlab}. The conversation's goal is usually modelled as a slot-filling problem. 
The \textit{dialogue manager} (DM) is the core component of these systems that chooses the dialogue actions according to the context. \textit{Reinforcement learning} (\textsc{RL}) can be used to model the DM, in which case the policy is trained to maximize the probability of satisfying the goal~\citep{gao2018neural}. 

We focus here on the multi-domain multi-task dialogue problem. In practice, real applications like personal assistants or chatbots must deal with multiple tasks: the user may first want to \textbf{find} a hotel (first task), then  \textbf{book} it (second task). Moreover, the tasks may cover several domains: the user may want to find a hotel (first task, first domain), book it (second task, first domain), and then find a restaurant nearby (first task, second domain). 

One way of handling this complexity is to rely on a \textit{domain hierarchy} which decomposes the decision-making process;
another way is to switch easily from one domain to another by scaling up the policy. 
Although \textit{structured dialogue policies} can adapt quickly from a domain to another \cite{chen2020structured}, covering multiple domains remains a hard task because 
it increases the dimensions of the state and action spaces while the reward signal remains sparse. A common technique to circumvent this reward scarcity 
is to guide the learning by injecting some knowledge through a teacher policy\footnote{\label{hc} For deployment the teacher is expected to be a human expert, however, for experimentation purposes we used the handcrafted policy as a proxy~\citep{casanueva_benchmarking_2017}.}.

Our main contribution is to study how structured policies like \textit{graph neural networks} (\textsc{GNN}) combined with some degree of \textit{imitation learning} (\textsc{IL}) can be effective to handle multi-domain dialogues.
We provide large scale experiments in a dedicated framework~\citep{zhu2020convlab} in which we analyze the performance of different types of policies, from multi-domain policy to generic policy, with 
different levels of imitation learning.

The remainder of this paper is structured as follows. We present the related work in Section~\ref{s:relwork}. Section~\ref{s:contr} presents our structured policies combined with imitation learning. The experiments and evaluation are described in Sections~\ref{s:expe} and~\ref{s:eval} respectively. Finally, we conclude in Section~\ref{s:concl}.



\section{Related Work}
\label{s:relwork}


Fundamental hierarchical reinforcement learning ~\cite{dayan1993feudal,parr1998reinforcement,sutton1999between,dietterich2000hierarchical} has inspired a previous string of works on dialogue management \cite{budzianowski2017sub, casanueva2018feudal, casanueva2018feudal2, chen2020structured}.
Recently, the use of structured hierarchy with \textsc{GNN} \citep{zhou2020graph, wu2020comprehensive} rather than a set of classical \textit{feed-forward networks} (\textsc{FNN}) enables the learning of non-independent sub-policies 
\citep{chen2018structured, chen2020distributed}.
These works adopted the \textit{Domain Independent Parametrisation} (\textsc{DIP}) that standardizes the slots representation into a common feature space to eliminate the domain dependence. It allows policies to deal with different slots in the same way. It is therefore possible to build 
policies that handle a variable number of slots and that transfer to different domains on similar tasks \citep{wang2015learning}. 

Our contribution differs from \citet{chen2020structured} on three points: first we perform our experiments on \textsc{ConvLab}~\cite{zhu2020convlab} which is a dedicated multi-domain framework; second, the \textit{dialogue state tracker} (\textsc{DST}) output is not discarded when activating the domain; third, we adapt the \textsc{GNN} structure to each domain by keeping the relevant nodes while sharing the edge's weights. 


The reward sparsity can be bypassed by guiding the learning through the injection of some knowledge via a teacher policy. This approach, called \textit{imitation learning} (\textsc{IL}) \citep{hussein2017imitation}, can be declined 
from pure \textit{behaviour cloning} (\textsc{BC}) where the agent only learns to mimic its teacher to pure \textit{reinforcement learning} (\textsc{RL}) where no hint is provided~\citep{shah2016interactive, hester2018deep, gordon2020learning, cordier2020diluted}. 

\begin{figure*}[ht!]
    \begin{center}
        \subfloat[\scriptsize{Pure \textit{ACER}}]{
        \includegraphics[width=0.43\textwidth]{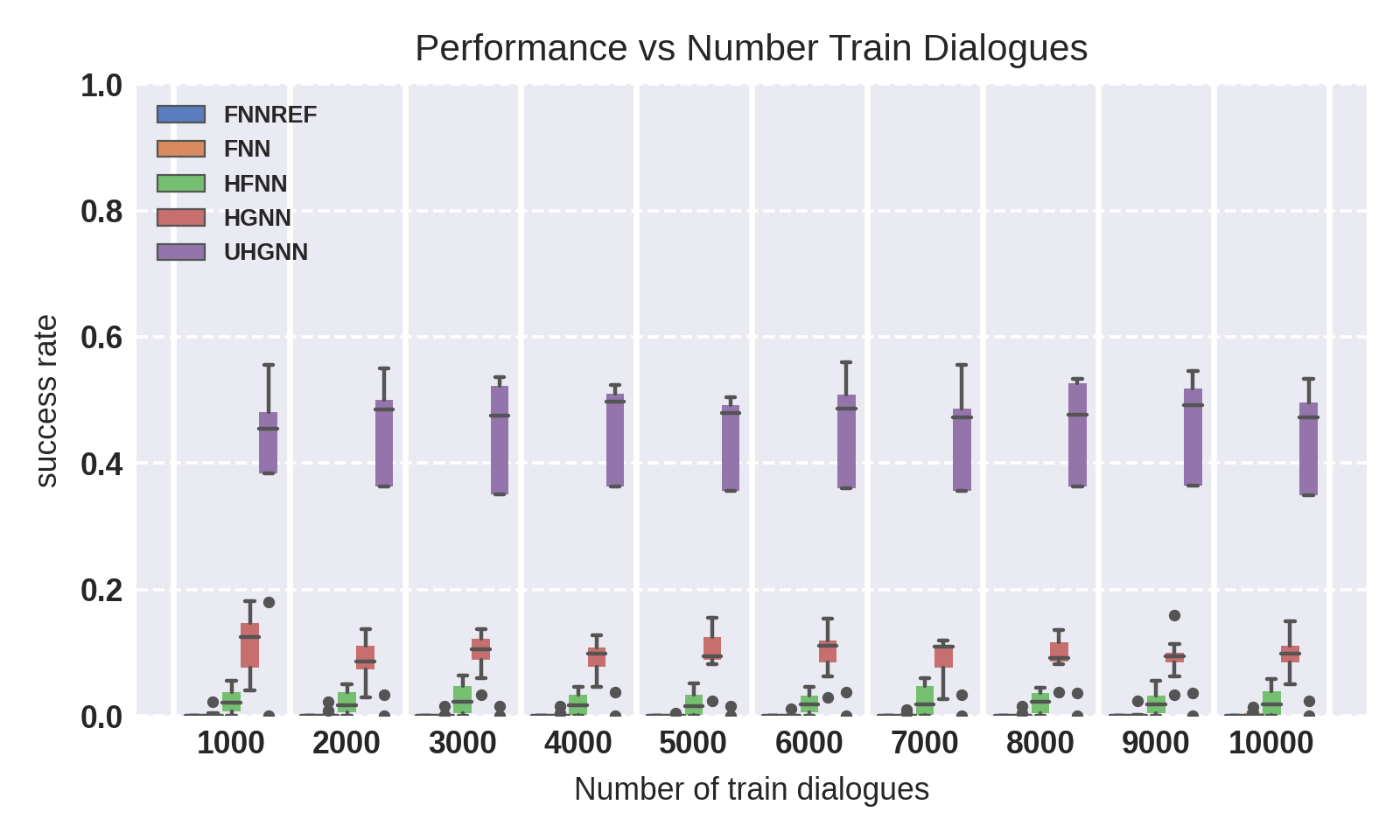}
        \label{subfig:convlab_a2c}
        }
        \subfloat[\scriptsize{Pure \textit{BC}}]{
        \includegraphics[width=0.43\textwidth]{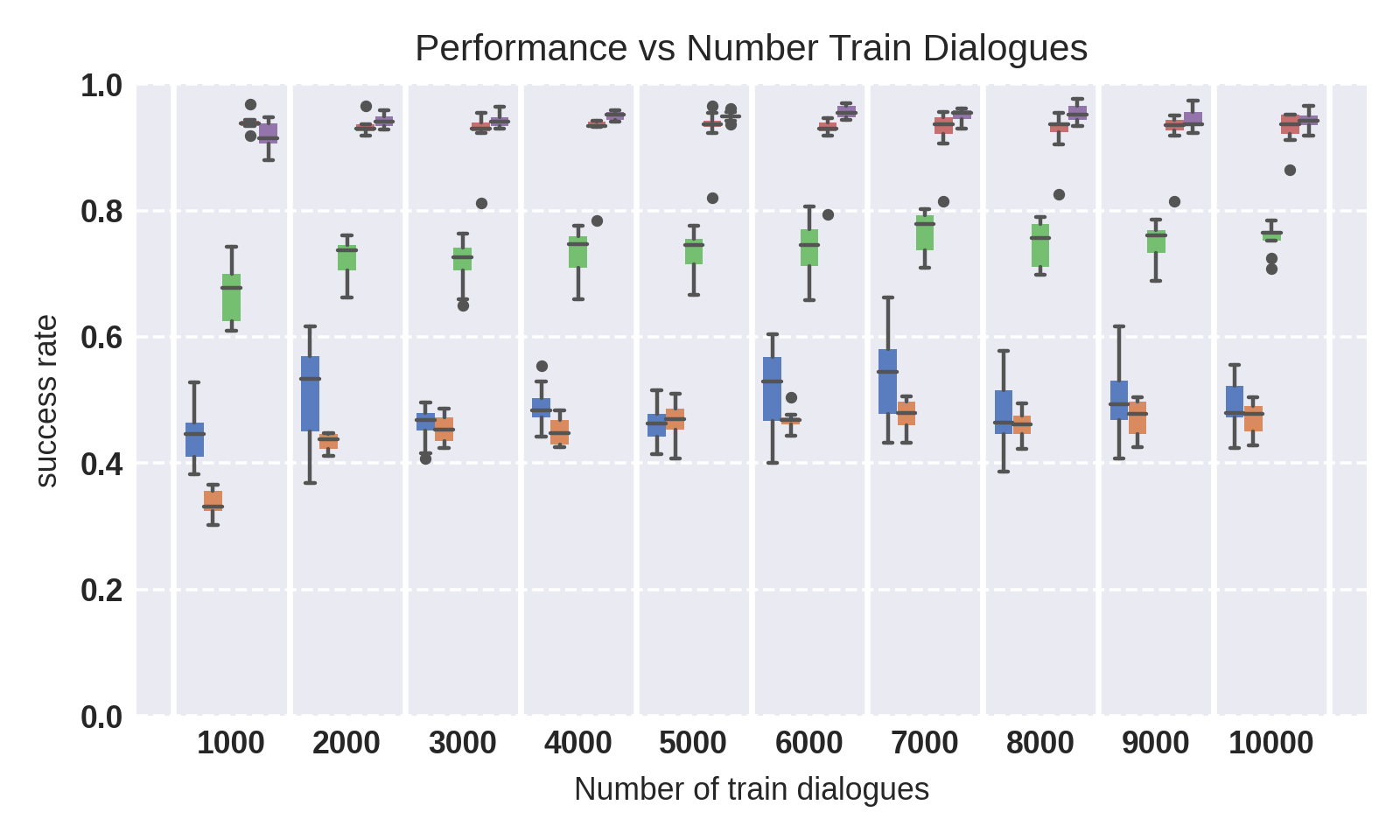}
        \label{subfig:convlab_bc}
        }\\
        \subfloat[\scriptsize{\textit{ACER} with \textit{ILfOD}.}]{
        \includegraphics[width=0.43\textwidth]{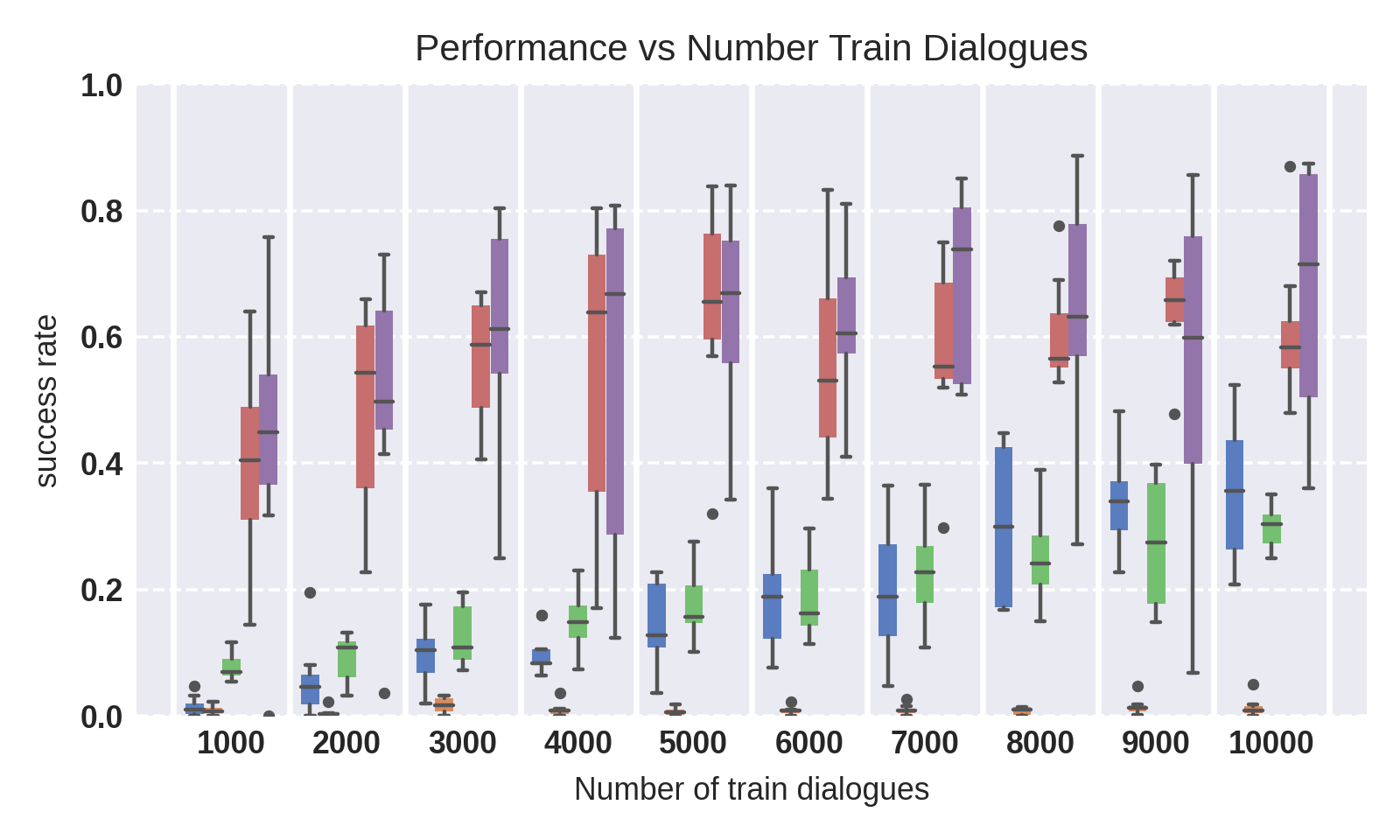}
        \label{subfig:convlab_a2cilfod}
        }
        \subfloat[\scriptsize{\textit{ACER} with \textit{ILfOS}}.]{
        \includegraphics[width=0.43\textwidth]{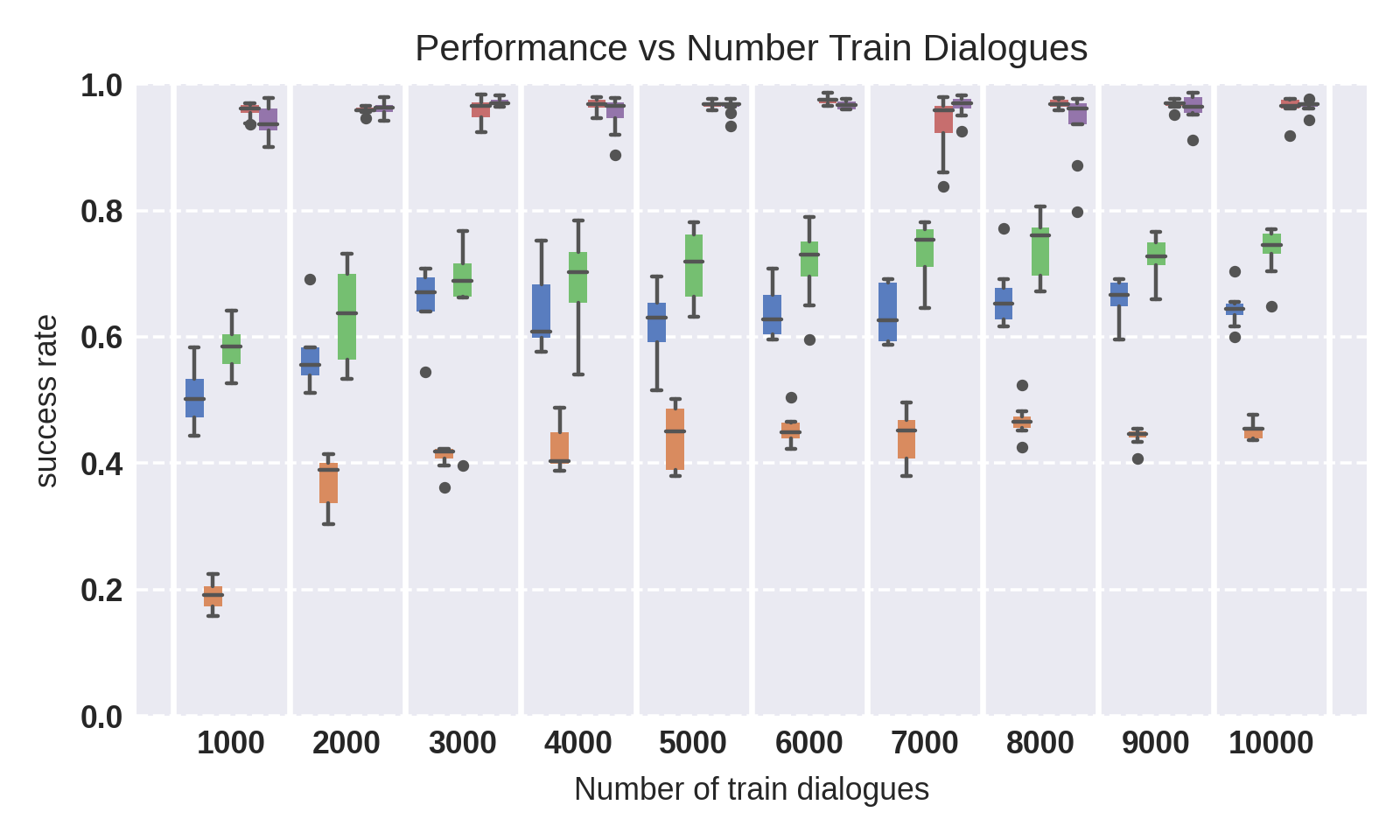}
        \label{subfig:convlab_a2cilfos}
        }\\
    \end{center}
    \caption{Distribution via boxplot of the performance of the proposed approaches on \textsc{ConvLab}, with 10 different initializations and without pre-training. The coloured area represents the interquartile Q1-Q3 of the distribution, the middle line represents its median (Q2) and the points are outliers. 
    }
    \label{fig:convlab_approach}
\end{figure*}



%


\section{Extended \textsc{GNN} Policies with Imitation} 
\label{s:contr}
We adopt the multi-task setting as presented in \textsc{ConvLab}, in which a single dialogue can have the following tasks: (i) \textbf{find}, in which the system requests information in order to query a database and make an offer; (ii) \textbf{book}, in which the system requests information in order to book the item. A single dialogue can also contain multiple domains such as \textit{hotel}, \textit{restaurant}, \textit{attraction}, \textit{train}, etc. 

Our method, illustrated in Figure~\ref{fig:proposal}, is designed to adapt: (i) at the domain-level (\textit{i.e.} be scalable to changes in the number of slots), and (ii) at the multi-domain-level (\textit{i.e.} be scalable to changes of domain). 
For each dialogue turn, it works as follow:
%
first, the \textsc{DST} module chooses which domain to activate. Then, the multi-domain belief state (and action space) is projected into the active domain (\textit{i.e} only the DIP nodes corresponding to the active domain are kept) as shown in Figure~\ref{sub:module_selection}. 
Afterwards, we apply the \textsc{GNN} message passing as \citet{chen2020structured} but only among the domain specific \textsc{DIP} nodes in the decision making module (Figure~\ref{sub:module_decision}).

\paragraph{GNN Policies}

The \textsc{GNN} structure we consider is a fully connected graph in which the nodes are extracted from the \textsc{DIP}. 
We distinguish two types of nodes: the slot nodes representing the parametrisation of each slot (denoted as \textsc{S-node}) and the general node representing the parametrisation of the domain (as \textsc{I-node} for {slot-Independent node}). This yields three types of edges: \textsc{I2S} (for \textsc{I-node} to \textsc{S-node}), \textsc{S2I} and \textsc{S2S}. 
This abstract structure is a way of modelling the relations between slots as well as exploiting symmetries based on weight sharing (Figure \ref{sub:module_decision}).

\paragraph{Imitation Learning}

In addition to the structured architecture, we use some level of \textsc{IL} to guide the agent's exploration.
In our experiments, we used \textsc{ConvLab}'s {handcrafted policy} as a \textit{teacher} (or \textit{oracle})\footnoteref{hc}, but other policies could be used as well.
\textit{Behaviour cloning} (\textsc{BC}) is a {pure supervised learning} method that tries to mimic the teacher policy. Its loss function is the cross-entropy loss as in a classification problem.
\textit{Imitation Learning From Oracle Demonstrations} (\textsc{ILfOD}) is a \textsc{RL} method which allows the agent to play oracle actions as demonstrations and to inject them in its \textit{replay buffer}. 
In our experiments, we kept half of the agent's own actions in the buffer along with those generated by the oracle.
\textit{Imitation Learning From Oracle Supervision} (\textsc{ILfOS}) is the combination of supervised and reinforcement learning
when the agent learns with a supervised loss, namely the margin loss \cite{hester2018deep}.

\begin{table*}[!htb]
\centering
\resizebox{1.99\columnwidth}{!}{%
\begin{tabular}{ccc|ccccccccc}
    \toprule
    \multicolumn{3}{c|}{\textbf{Configuration}} & \textbf{Avg Turn} & \textbf{Inform (\%)} & \textbf{Book} & \multicolumn{2}{c}{\textbf{Complete}} & \multicolumn{2}{c}{\textbf{Success}} \\
    \textbf{NLU} & \textbf{Policy} & \textbf{NLG} &  \textbf{(succ/all)} & \textbf{Prec. / Rec. / F1} & \textbf{Rate (\%)} & \multicolumn{2}{c}{\textbf{Rate (\%)}} & \multicolumn{2}{c}{\textbf{Rate (\%)}} \\
    \midrule \midrule
    -  & HDC & - & 10.6/10.6 & 87.2 / 98.6 / 90.9 & 98.6 & 97.9 & - & 97.3 & - \\
    -     & ACGOS (ours) & -        & 13.1/13.2 & 94.8 / 99.0 / 96.1 & 98.7 & 98.2 & (+0.3) & 97.0 & (-0.3) \\ 
    \midrule
    \textsc{BERT}  & HDC & T & 11.4/12.0 & 82.8 / 94.1 / 86.2 & 91.5 & 92.7 & - & 83.8 & - \\
    \textsc{BERT}  & HDC$^{\dagger}$ & T & 11.6/12.3 & 79.7 / 92.6 / 83.5 & 91.1 & 90.5 & (-2.2) & 81.3 & (-2.5) \\
    \textsc{BERT}  & \textsc{MLE}$^{\dagger}$  & T & 12.1/24.1 & 62.8 / 69.8 / 62.9 & 17.6 & 42.7 & (-50.0) & 35.9 & (-47.9) \\
    \textsc{BERT}  & \textsc{PG}$^{\dagger}$   & T & 11.0/25.3 & 57.4 / 63.7 / 56.9 & 17.4 & 37.4 & (-55.3) & 31.7 & (-52.1) \\
    \textsc{BERT}  & \textsc{GDPL}$^{\dagger}$ & T & 11.5/21.3 & 64.5 / 73.8 / 65.6 & 20.1 & 49.4 & (-43.3) & 38.4 & (-45.4) \\
    \textsc{BERT}  & \textsc{PPO}$^{\dagger}$  & T & 13.1/17.8 & 69.4 / 85.8 / 74.1 & 86.6 & 75.5 & (-17.2) & 71.7 & (-12.1) \\
    \textsc{BERT}  & \textsc{ACGOS} (ours) & T & 14.0/14.8 & 88.8 / 92.6 / 89.5 & 86.6 & 89.1 & (-3.6) & 81.7 & (-2.1) \\
    \bottomrule
\end{tabular}
}
\caption{Dialogue system evaluation with simulated users. T means template-based \textsc{NLG}. Configurations without \textsc{NLU} and \textsc{NLG} modules pass directly the dialogue act. Configurations with \textsc{ACGOS} and HDC policies are evaluated on a single run with 1,000 dialogues. 
Configurations with ${\dagger}$ are taken from the \href{https://github.com/thu-coai/ConvLab-2/tree/ad32b76022fa29cbc2f24cbefbb855b60492985e}{GitHub of \textsc{ConvLab}}. \textsc{PPO} in \textsc{ConvLab} used behaviour cloning as the pre-trained weights (see for \href{https://github.com/thu-coai/ConvLab-2/issues/54 }{more details}). 
}
\label{tab:sotatableconvlab}
\end{table*}


\section{Experiments}
\label{s:expe}


We performed an ablation study: (i) by progressively extending the baseline to our proposed \textsc{GNN}s and (ii) by guiding the exploration with \textsc{IL}.
All the experiments were restarted 10 times with random initialisations and the results evaluated on 500 dialogues were averaged. Each learning trajectory was kept up to 10,000 dialogues with a step of 1,000 dialogues in order to analyse the variability and stability of the methods.


\paragraph{Models}

The baseline is \textsc{\textbf{ACER}} which is a sophisticated actor-critic method~\citep{wang2016sample}. After an ablation study, we progressively added some notion of hierarchy to \textsc{FNN}s to approximate the structure of \textsc{GNN}s. 
\textsc{\textbf{FNN}} is a feed-forward neural network  with \textsc{DIP} parametrisation. Thus, the agent actions are single-actions.
\textsc{\textbf{FNN-REF}} is a FNN with the native parametrisation (no \textsc{DIP}) with multiple-actions of \textsc{ConvLab}\footnote{The native parametrisation manually groups multi-actions based on \textsc{MultiWOZ}~\cite{budzianowski2018multiwoz}.}.
\textsc{\textbf{HFNN}} is a hierarchical policy with domain-selection module and based on \textsc{FNN}s for each domain. 
\textsc{\textbf{HGNN}} is a hierarchical policy with domain-selection module and based on \textsc{GNN}s. 
\textsc{\textbf{UHGNN}} is a \textsc{HGNN} with a unique \textsc{GNN} for all domains. 

\paragraph{Metrics}
We evaluate the performance of the policies for all tasks. For the {find} task, we use the precision, the recall and the F-score metrics: the \textbf{inform rates}. For the {book} task, we use the accuracy metric namely the \textbf{book rate}.
The dialogue is marked as \textbf{successful} if and only if both inform's recall and book rate are 1.
The dialogue is considered \textbf{completed} if it is successful from the user's point of view (\textit{i.e} a dialogue can be completed without being successful if the information provided is not the one objectively expected by the simulator).



\section{Evaluation}
\label{s:eval}
We evaluate the dialogue manager and the dialogue system both with simulated users.

\paragraph{Dialogue Manager }
\label{ss:evaldm}
We performed an ablation study based on \textsc{ACER} as reported in Figure~\ref{fig:convlab_approach}.
First, all \textsc{RL} variants of \textsc{ACER} (Figure~\ref{subfig:convlab_a2c}) have difficulties to learn without supervision in contrast to \textsc{BC} variants (Figure~\ref{subfig:convlab_bc}).
In particular, we see that hierarchical decision making networks (\textsc{HFNN} in green), graph neural network (\textsc{HGNN} in red) and generic policy (\textsc{UHGNN} in purple) drastically improve the performance compared to \textsc{FNN}s. Similarly, using \textsc{IL} like \textsc{ILfOD} (Figure~\ref{subfig:convlab_a2cilfod}) and \textsc{ILfOS} (Figure~\ref{subfig:convlab_a2cilfos}) notably improves the performance. 
Therefore, learning generic \textsc{GNN}s allows collaborative gradient update and efficient learning on multi-domain dialogues.
Conversely, we observe that hierarchical decision making 
with \textsc{HFNN}s does not systematically guarantee any improvement. 
These results suggest that \textsc{GNNs} are useful for learning dialogue policies on multi-domain 
which can be transferred during learning across domains on-the-fly to improve performance.
Finally, regarding \textsc{ILfOD} variants (Figure~\ref{subfig:convlab_a2cilfod}), we can observe that all architectures are affected by a large variability. This shows that multi-domain dialogue management is difficult despite the use of demonstrations and that learning with reward is not sufficient to robustly succeed.



\paragraph{Dialogue System }

We evaluate the policy learning algorithms in the entire dialogue pipeline, in particular our best \textsc{DM} policy \textsc{ACER-ILfOS-UHGNN} under a shorter name \textsc{\textbf{ACGOS}}. 
The results of our experimentation are presented in Table~\ref{tab:sotatableconvlab}.
We observe that the performance of our approach is closed to the handcrafted policy (the teacher) when directly passing the dialogue acts (97.3 \textit{vs.} 97.0). It is also closed to the handcrafted policy when using \textsc{BERT NLU} \cite{devlin2018bert} and template-based \textsc{NLG} (83.8 \textit{vs.} 81.7). It is much better compared to the baselines with a significant difference (\textit{e.g.} with 81.7 for \textsc{ACGOS} \textit{vs.} 71.7 for pre-trained \textsc{PPO}).
These results highlight the benefit of structured policies against standard policies.


\section{Conclusion}
\label{s:concl}


We studied structured policies like \textsc{GNN} combined with some imitation learning 
that effectively handle multi-domain dialogues.
The results of our large-scale experiments on \textsc{ConvLab} confirm that an actor-critic based policy with a \textsc{GNN} structure can solve multi-domain multi-task dialogue problems.
Finally, we evaluated our best policy (\textsc{ACGOS}) in a complete dialogue system with simulated users. 
It overcomes the baselines and it is comparable to the handcrafted policy. 

A limitation of current policies in~\textsc{ConvLab}, including ours, is that the robustness to noisy inputs is not specifically addressed as it had been done in PyDial~\cite{ultes2017pydial}.
It could be also interesting to study the impact of incorporating real human feed-backs and demonstrations instead of a handcrafted teacher. 

The \textsc{GNN} structured policies combined with imitation learning avoid sparsity, while being data efficient, stable and adaptable. They are relevant for covering multi-domain task dialogue problems.






\nocite{}

\bibliography{anthology,custom}
\bibliographystyle{acl_natbib}


\newpage

\appendix

\section{Appendix}
\label{sec:appendix}

\subsection{Domains}

\begin{table}[!ht]
	\resizebox{0.95\columnwidth}{!}{%
	\begin{tabular}{lcc}
		\toprule
		Domain & \# constraint slots & \# request slots \\
		\midrule
		\midrule
		\textsc{\textbf{ConvLab}} & find/book & search \\
		\midrule
		Restaurant & 4/3 & 5 
		\\
		Attraction  & 3/- & 7 
		\\
		Hotel       & 7/3 & 5 
		\\
		Taxi        & 4/- & 2 \\
		Train       & 5/1 & 5 \\
		Hospital    & 1/- & 3 \\
		Police      & -/- & 3 \\
		\bottomrule
	\end{tabular}
	}
	\caption{Domains Description of \textsc{ConvLab} framework}
	\label{tab:domains_convlab}
\end{table}

\begin{figure}[ht!]
  \begin{center}
    \includegraphics[width=0.9\columnwidth]{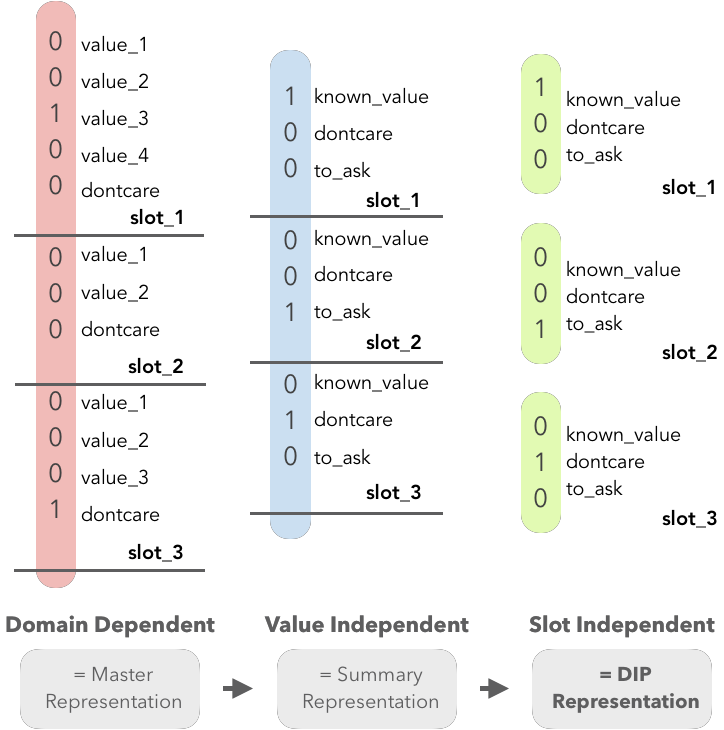}
    \caption{Transformation from initial state to DIP state representation (it works similarly for actions).}
    \label{fig:state_action}
  \end{center}
\end{figure}

\paragraph{Belief State} The belief state representation is deterministic. As shown in Figure \ref{fig:state_action}, there is no uncertainty (all values are either 0's or 1's). 

\paragraph{State Space} The input to the dialogue manager is the belief state which is a dictionary of all tractable information (slot-value pairs, history, dialogue actions of system and user, etc.). This is called the \textit{master state space}. And, due to its large size, the representation is projected into the \textit{summary state space} by a process called \textit{value abstraction} \cite{wang2015learning}. Finally, it must be vectorised in order to be interpretable by neural networks. 

\paragraph{Action Space} The dialogue manager's output is a probabilistic distribution over all possible actions. To reduce the complexity of the learning problem, \textit{master actions}, which are valued dialogue acts such as \textsc{inform}(date = ’2022-01-15’), are abstracted into \textit{summary actions} like \textsc{inform}(date), the \textit{value abstraction} module being in charge of restoring the relevant values in the context. On \textsc{ConvLab} the policy may activate several actions simultaneously (called \textit{multiple-actions}). 

\paragraph{Domain Independent Parametrisation} (or \textsc{DIP}) \citep{wang2015learning} standardises the slots representation into a common feature space to eliminate the domain dependence. In particular, the \textsc{DIP} state and action representations are not reduced to a flat vector but to a set of sub-vectors: one corresponding to the domain parametrisation (called \textit{slot-independent representation}), the others to the slots parametrisation (called \textit{slot-dependent representations}). 

\begin{table}[!hp]
	\centering
	\begin{tabular*}{\linewidth}{@{\extracolsep{\fill}}p{0.99\linewidth}}
		\toprule
		Component / Description \\
		\midrule
		\midrule
		\textbf{Beliefs} \\
		\midrule
		\textbf{constraint slot beliefs}: $\{b_{d,s}^{inf} \in \mathcal{V}_s,\, \forall s \in \mathcal{S}_d^{inf},\, \forall d \in \mathcal{D}\}$
		    The goal constraints belief for each informable slot. This is either an assignment of a value from the ontology which the user has specified as a constraint, or has a special value — either \textit{dontcare} which means the user has no preference, or \textit{none} which means the user is yet to specify a valid goal for this slot. To be exact, for each domain, the constraint slot dictionary separates slots with respect to the task i.e we distinguish the \textit{find} slot dictionary and the \textit{book} slot dictionary. \\
		\textbf{request slot beliefs}: $\{b_{d,s}^{req} \in \mathbb{B},\, \forall s \in \mathcal{S}_d^{req},\, \forall d \in \mathcal{D}\}$:
		    A set of requested slots, i.e. those slots whose values have been requested by the user, and should be informed by the system.\\
		\midrule
		\textbf{Features} \\
		\midrule
		\textbf{terminated}: $f_1 \in \mathbb{B}$:
		    A boolean showing that the user wants to end the call. \\
		\textbf{booked}: $f_2 \in \mathcal{V}_{DB(d)}$:
		    The name of the last venue offered by the system to the user with respect to the constraint slots with additional information like reference. To be exact, this feature is located in the \textit{book} slot dictionary.\\
		\textbf{degree pointer}: $f_3 \in\mathbb{B}^6$:
		    The vector counting the number of entities \textit{count} matching with constraint slots in acceptance list: [count==0, count==1, count==2, count==3, count==4, count>=5]. \\
		\midrule
		\textbf{System Acts} \\
		\midrule
		\textbf{system acts}: $a^{sys} \in list(\mathcal{A}^{sys})$: The list of the last system actions. \\
		\midrule
		\textbf{User Acts} \\
		\midrule
		\textbf{user acts}: $a^{user} \in list(\mathcal{A}^{user})$: The list of the last user actions. \\
		\bottomrule
	\end{tabular*}
	\caption{Belief State Template in \textsc{ConvLab} framework}
	\label{tab:belief_state_convlab}
\end{table}

\subsection{State and Action Representations}

We propose to formally present the state representations used in our experiments. For details about our notations, see Table~\ref{tab:belief_state_convlab}.


\begin{subequations}
\textbf{Flat state representation in \textsc{ConvLab}}
\begin{multline*}
    \phi(x) = \Big(\bigoplus_{s \in \mathcal{S}^{inf}} b_s^{inf}\Big) \\
    \oplus a^{user} \oplus a^{sys} \oplus [f_1] \oplus f_2 \oplus f_3
\end{multline*}
\end{subequations}
where $x$ is the initial state, $\phi(x)$ is the full state parametrisation, $\mathcal{S}^{inf}$ is the set of informable slots, $b_s^{inf}$ is the one-encoding vector of the informable slot $s$, $a^{user}$ and $a^{sys}$ are the one-encoding vectors of previous user and system actions, $f_1$ is the boolean "terminated dialogue", $f_2$ is the boolean "booked offer" with respect to each domain, $f_3$ is the one-encoding vector of the matching entities count with respect to each domain and $\oplus$ is the vector concatenation operator.


\begin{subequations}

\textbf{\textsc{DIP} state representation}

Slot independent parametrisation:
\begin{equation*}
    \phi_d(x) = a^{user}|_g \oplus a^{sys}|_g \oplus [f_1,\, f_2|_d,\, f_3|_d]
\end{equation*}
where $x$ is the initial state, $\phi_d(x)$ is the active domain state parametrisation, $a^{user}|_g$ and $a^{sys}|_g$ are the one-encoding vectors of previous general user and system actions, $f_1$ is the boolean "terminated dialogue", $f_2|_d$ is the boolean "booked offer" with respect to the active domain, $f_3|_d$ is the one-encoding vector of the matching entities count with respect to the active domain and $\oplus$ is the vector concatenation operator.

Slot dependent parametrisation:
\begin{align}
    &\forall s_i \in \mathcal{S}_d,\, \phi_{s_i}(x) =
    a^{user}|_{s_i} \oplus a^{sys}|_{s_i} \notag\\ 
    &\qquad \oplus \big[\mathbbm{1}(\exists\, v \in \mathcal{V}_{s_i}/\{\textrm{none}\},\, b_{s_i}^*[v]=1)\big] \label{ml:l1}\\
    &\qquad \oplus \big[\mathbbm{1}(s_i \in \mathcal{S}_d^{inf})\big] \label{ml:l2}\\ 
    &\qquad \oplus \big[\mathbbm{1}(s_i \in \mathcal{S}_d^{req})\big] \label{ml:l3}
\end{align}
\end{subequations}
where $x$ is the initial state, $\phi_{s_i}(x)$ is the slot parametrisation of the $i^{th}$ slot, $\mathcal{S}_d$ is the set of slots of the active domain, $a^{user}|_{s_i}$ and $a^{sys}|_{s_i}$ are the one-encoding vectors of previous user and system actions of the $i^{th}$ slot, (\ref{ml:l1}) is the indicator of known value, (\ref{ml:l2}) is the indicator of informable slot and (\ref{ml:l3}) is the indicator of requestable slot and $\oplus$ is the vector concatenation operator.

\subsection{Implementation Details}


\paragraph{Imitation learning}

The used oracle is the handcrafted agent proposed by each framework.
When we use \textsc{ILfOD} or \textsc{ILfOS} methods, $50\%$ of the time the oracle trajectories is used.
When we use \textsc{ILfOS}, we call also in $100\%$ of the time the oracle which gives us the best expert action as supervision and a margin penalty $\mu = \log(2)$ \cite{hester2018deep}.

\paragraph{Reinforcement learning}

Our policy algorithm is an off-policy learning that uses experience replay (all data are stored in buffers) without priority \textit{i.e} without importance sampling.
The exploitation-exploration procedure 
is achieved by Boltzmann sampling with a fixed temperature $\tau=1$.

\paragraph{Metrics and Rewards}

\textbf{Inform recall} evaluates whether all the requested information has been informed when \textbf{inform precision} evaluates whether only the requested information has been informed.
\textbf{Book rate} assesses whether the offered entity meets all the constraints specified in the user goal.
The system is guided by the rewards as follows. If all domains are solved (a domain is solved if all related tasks are solved), it gains $40$ points. If the current active domain is solved, it gains $5$ points. Otherwise, it is penalised by $1$ point.

\paragraph{Model setup for neural network architectures}

Our \textsc{FNN} models have two hidden layers, both with $128$ neurons. 
Our \textsc{GNN} models have one first hidden layer with $32$ neurons for each node (two in all: \textsc{S-node} and \textsc{I-node}). Then the second hidden layer is composed of $32$ neurons for each relation (three in all: \textsc{S2S}, \textsc{S2I} and \textsc{I2S}).
The size of the tested networks are of the order of magnitude of $10\,000$ to more than $100\,000$ parameters.


For learning stage, we use a learning rate $\mathrm{lr}~=~10^{-3}$, a dropout rate $\mathrm{dr}~=~0.1$ and a batch size $\mathrm{bs}~=~64$. Each loss function has a weight of $\lambda_Q=0.5$, $\lambda_\pi=1.$, $\lambda_{IL}=1.$ and $\lambda_{ent}=0.01$ respectively. 
The learning frequency is one iteration after each episode (finished dialogue) with only one gradient iteration.

\paragraph{Used packages for the experiment}
We used the dialogue system frameworks named \textsc{ConvLab}~\citep{zhu2020convlab}. For the implementation of neural networks, we used \textsc{Pytorch}~\citep{NEURIPS2019_9015} in our dialogue systems. We also used another toolkit for reinforcement learning research named \textsc{OpenAI Gym}~\citep{brockman2016openai}.




\subsection{Supplementary Results}

\begin{figure}[ht!]
  \begin{center}
    \subfloat[Recall Average - \textsc{UHGNN} models]{
        \includegraphics[width=0.4\textwidth]{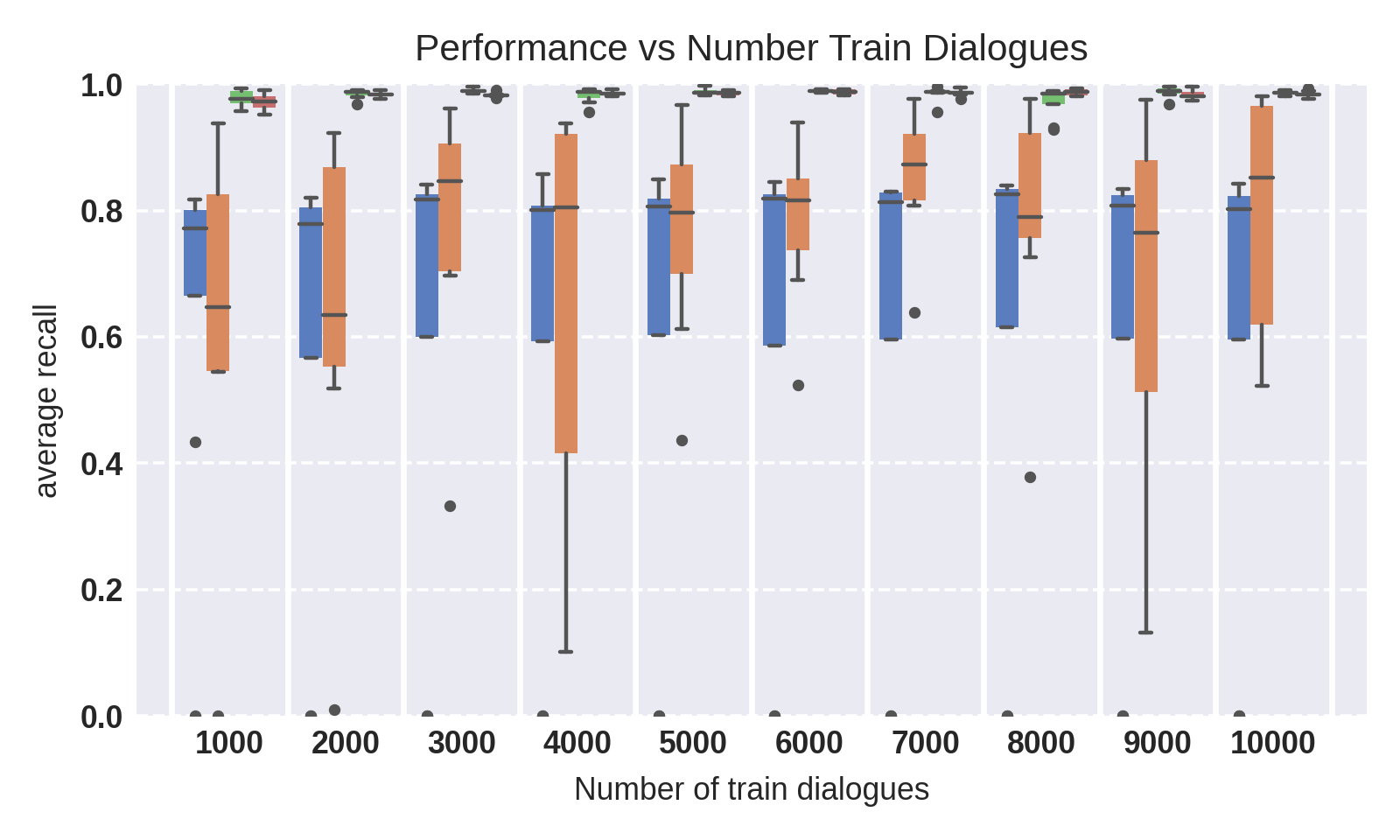}
        \label{sub:recall_UHGNN}
    }\\
    \subfloat[Recall Average - \textsc{HGNN} models]{
        \includegraphics[width=0.4\textwidth]{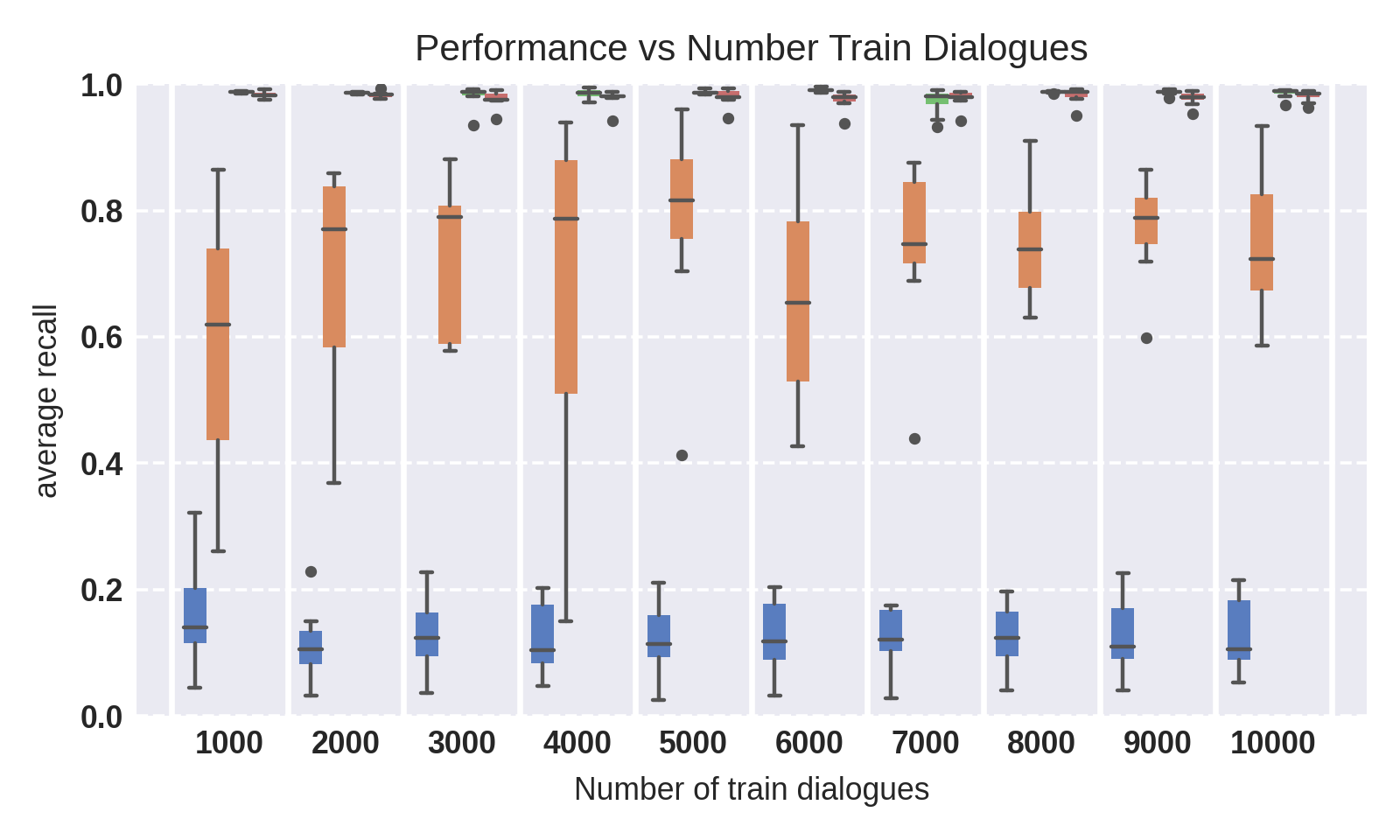}
        \label{sub:recall_HGNN}
    }\\
    \subfloat[Recall Average - \textsc{HFNN} models]{
        \includegraphics[width=0.4\textwidth]{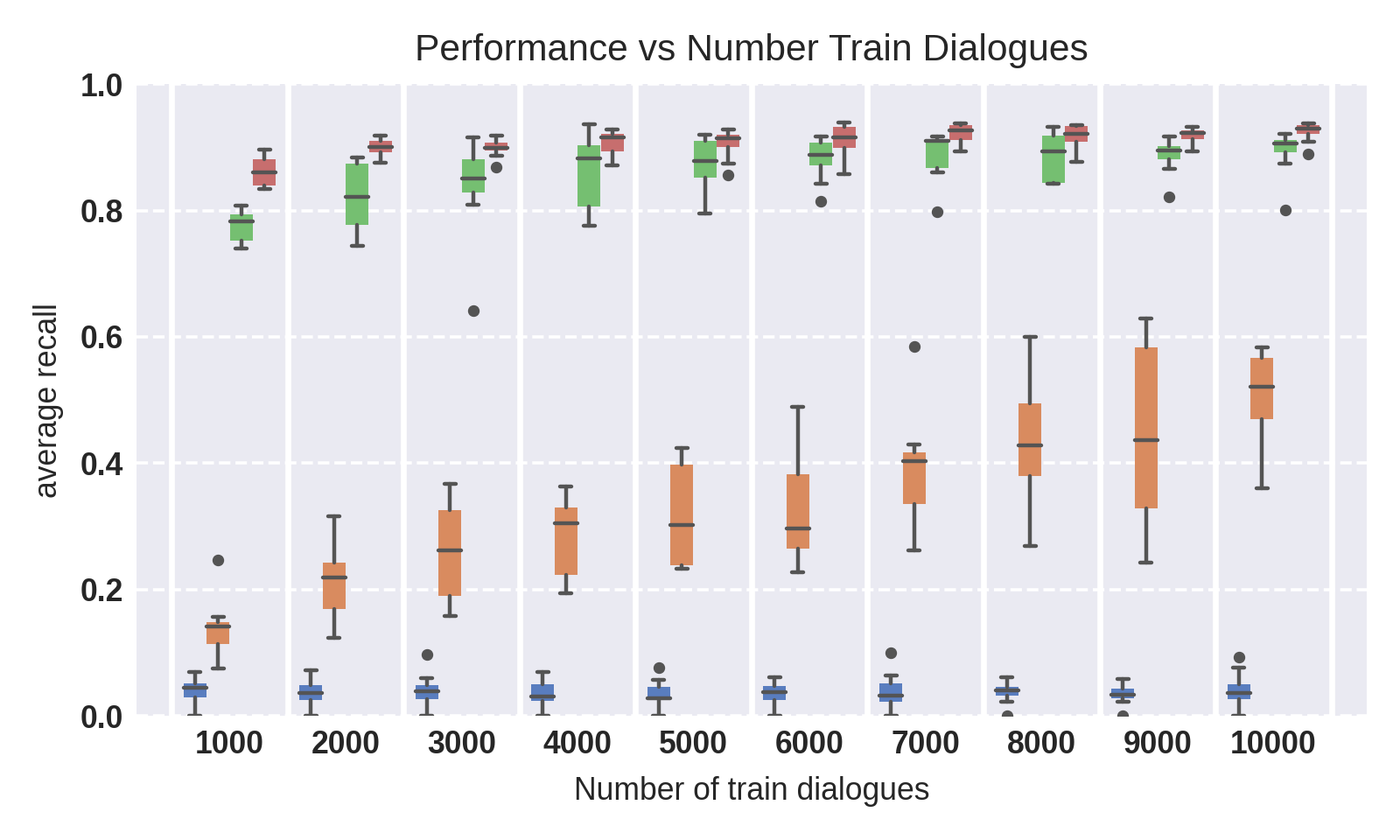}
        \label{sub:recall_HFNN}
    }\\
    \subfloat[Recall Average - \textsc{FNN} models with \textsc{DIP} parametrization]{
        \includegraphics[width=0.4\textwidth]{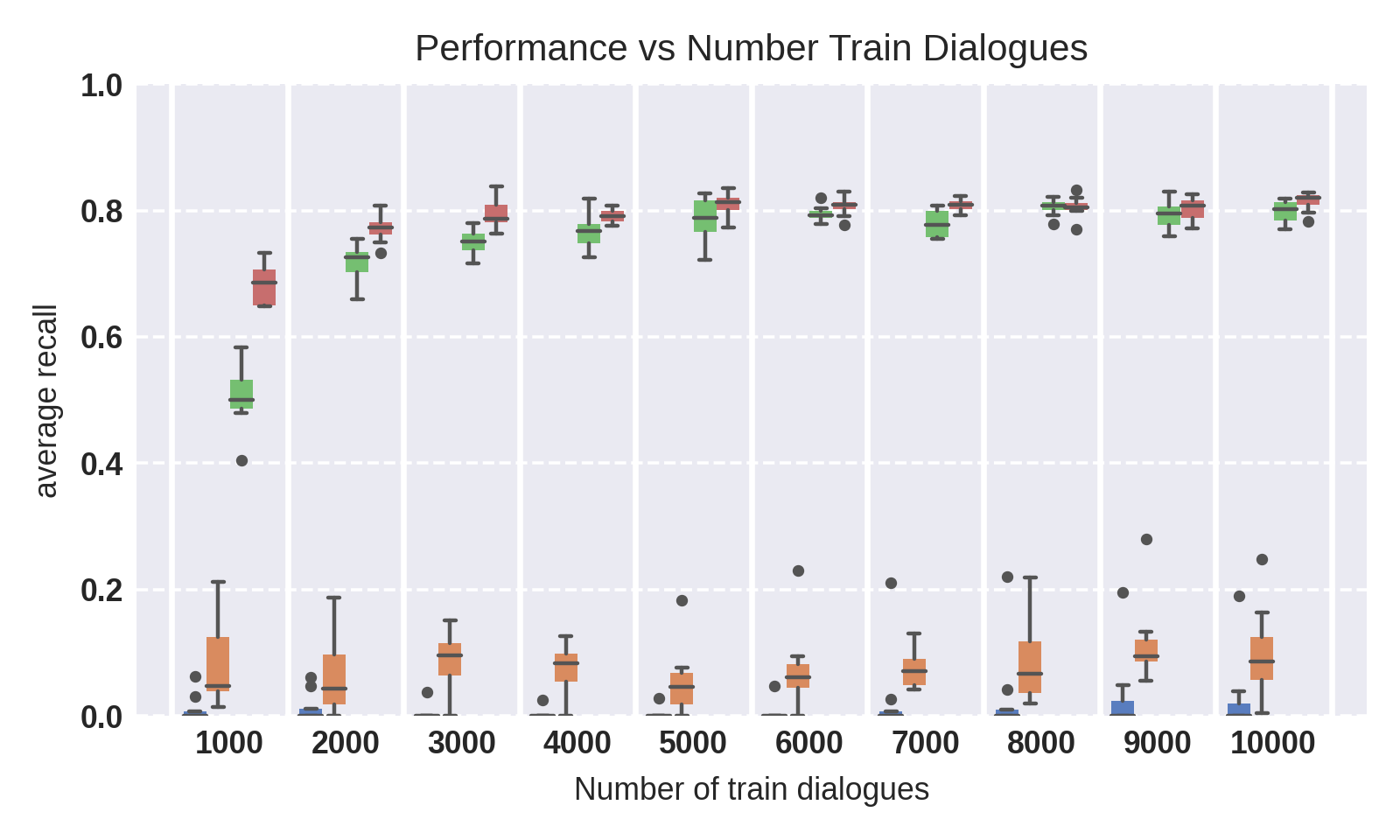}
        \label{sub:recall_FNN}
    }\\
    \subfloat[Recall Average - \textsc{FNN} models with native parametrization]{
        \includegraphics[width=0.4\textwidth]{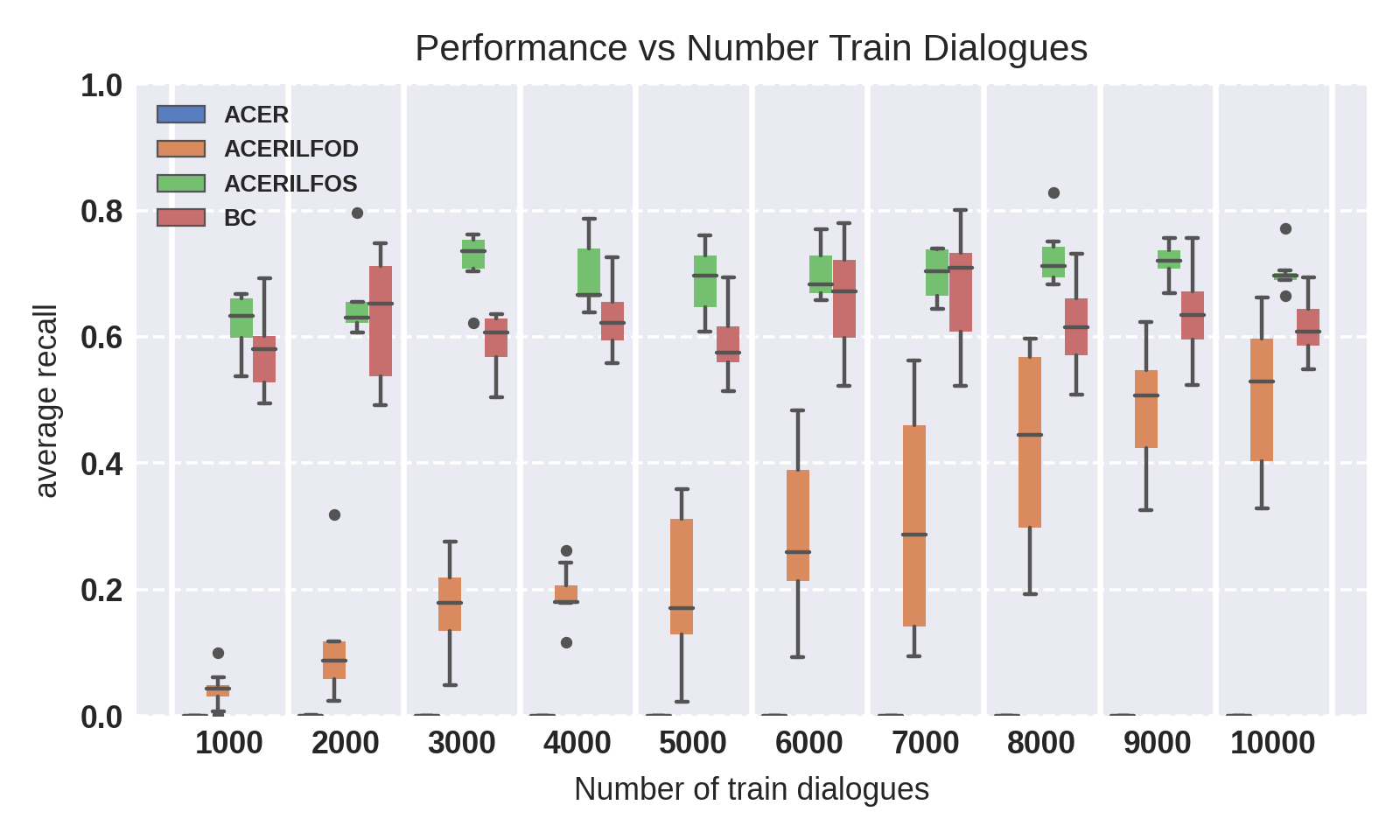}
        \label{sub:recall_FNN_REF}
    }\\
  \end{center}
  \caption{Summary of performance - Task \textit{find}}
  \label{fig:all_recall}
\end{figure}

\begin{figure}[ht!]
  \begin{center}
    \subfloat[Book Rate - \textsc{UHGNN} models]{
        \includegraphics[width=0.4\textwidth]{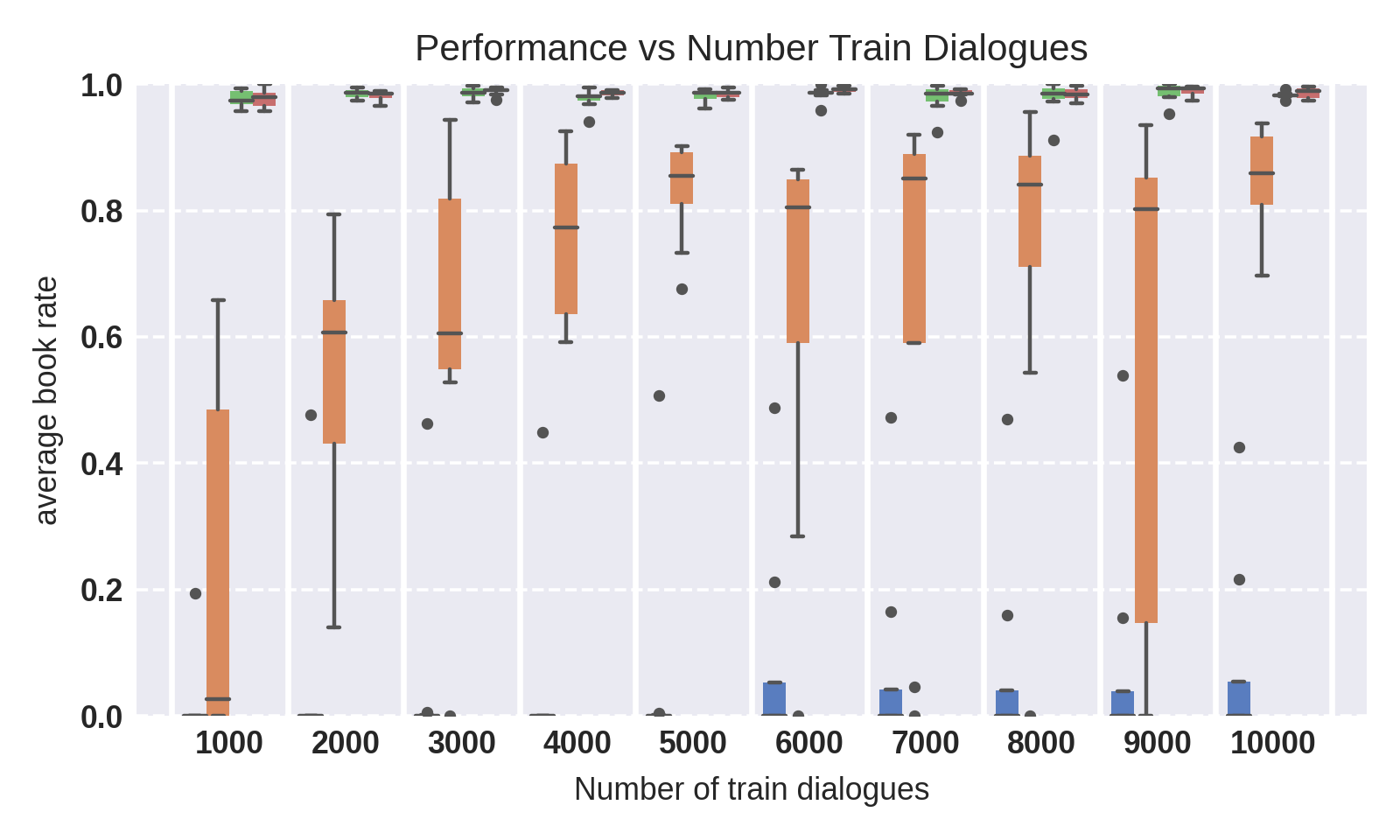}
        \label{sub:match_UHGNN}
    }\\
    \subfloat[Book Rate - \textsc{HGNN} models]{
        \includegraphics[width=0.4\textwidth]{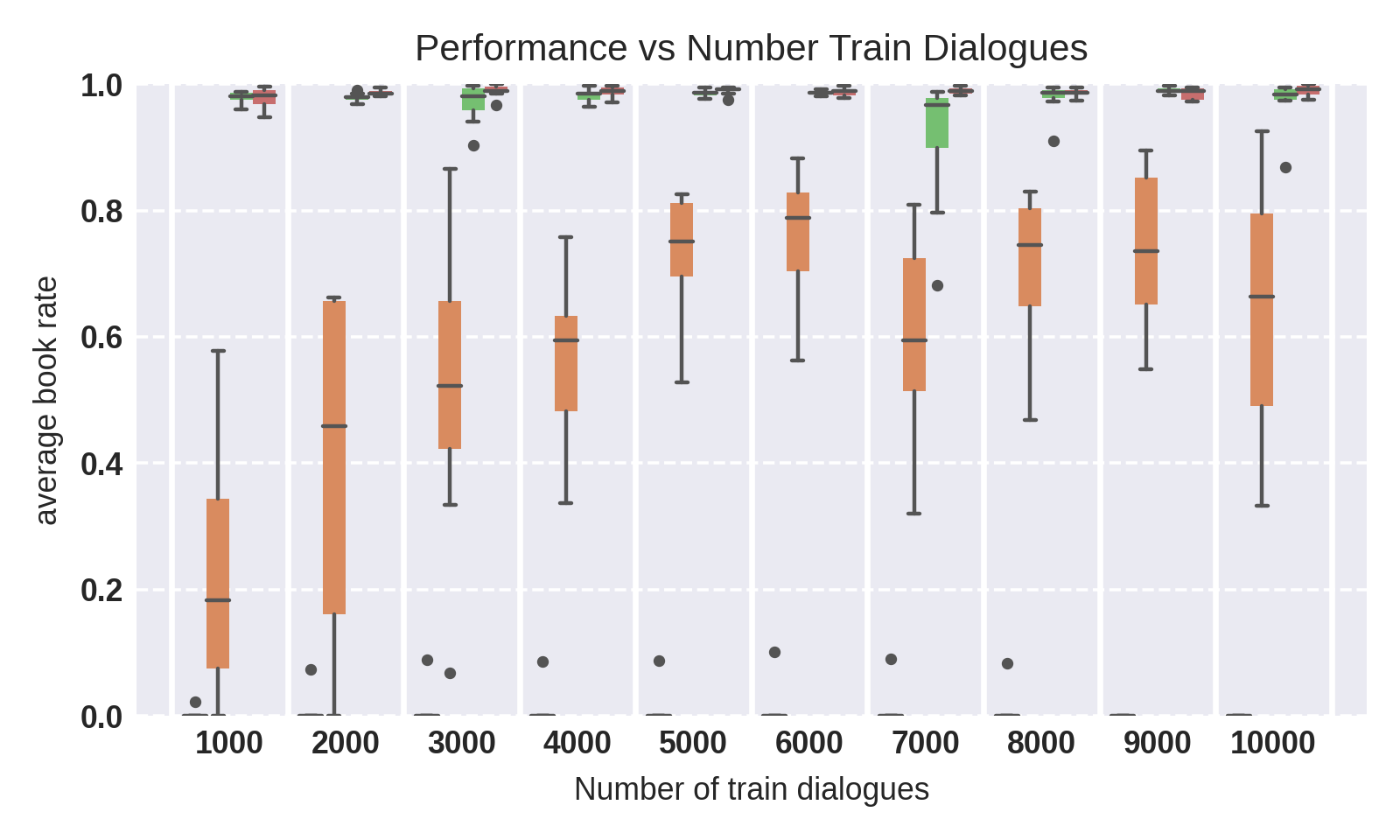}
        \label{sub:match_HGNN}
    }\\
    \subfloat[Book Rate - \textsc{HFNN} models]{
        \includegraphics[width=0.4\textwidth]{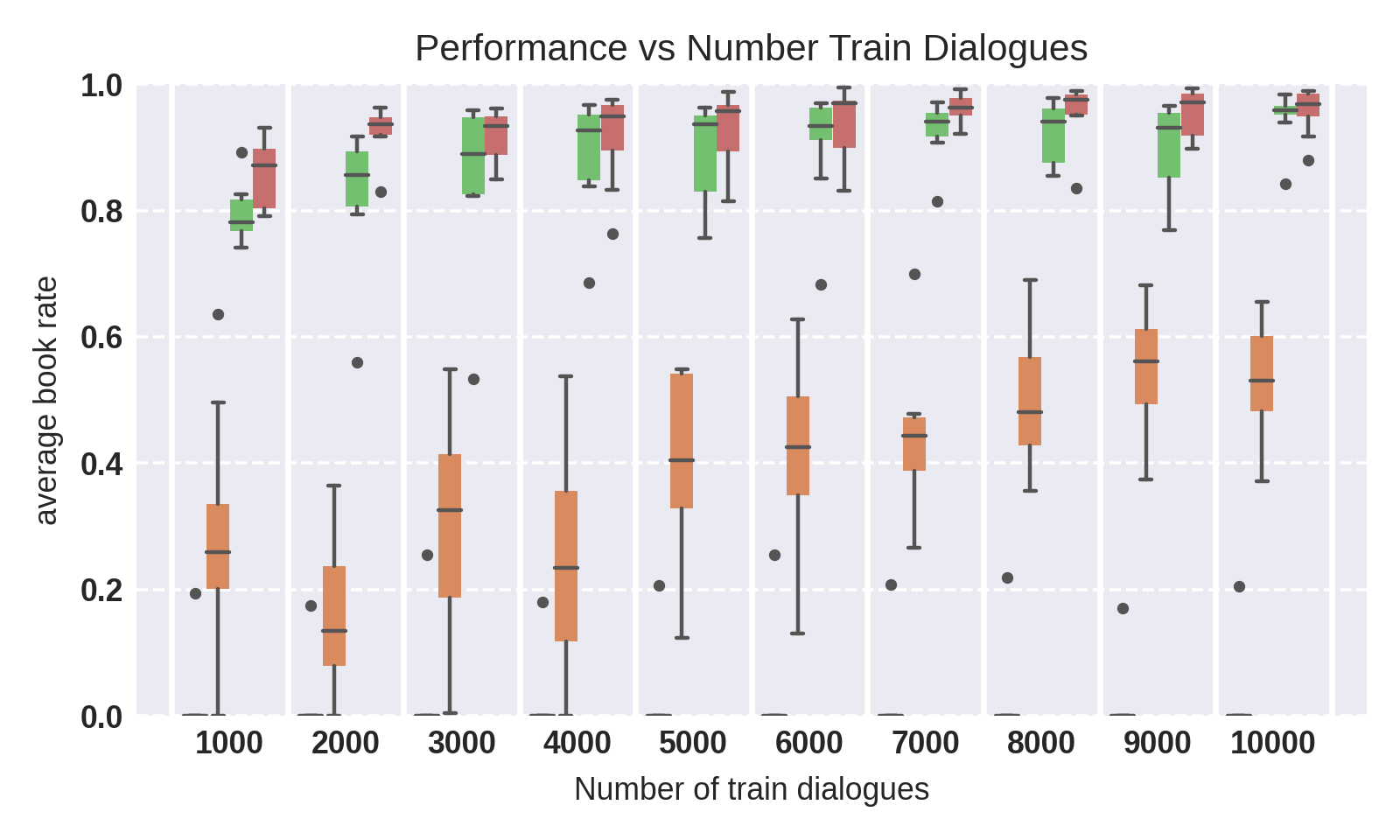}
        \label{sub:match_HFNN}
    }\\
    \subfloat[Book Rate - \textsc{FNN} models with \textsc{DIP} parametrization]{
        \includegraphics[width=0.4\textwidth]{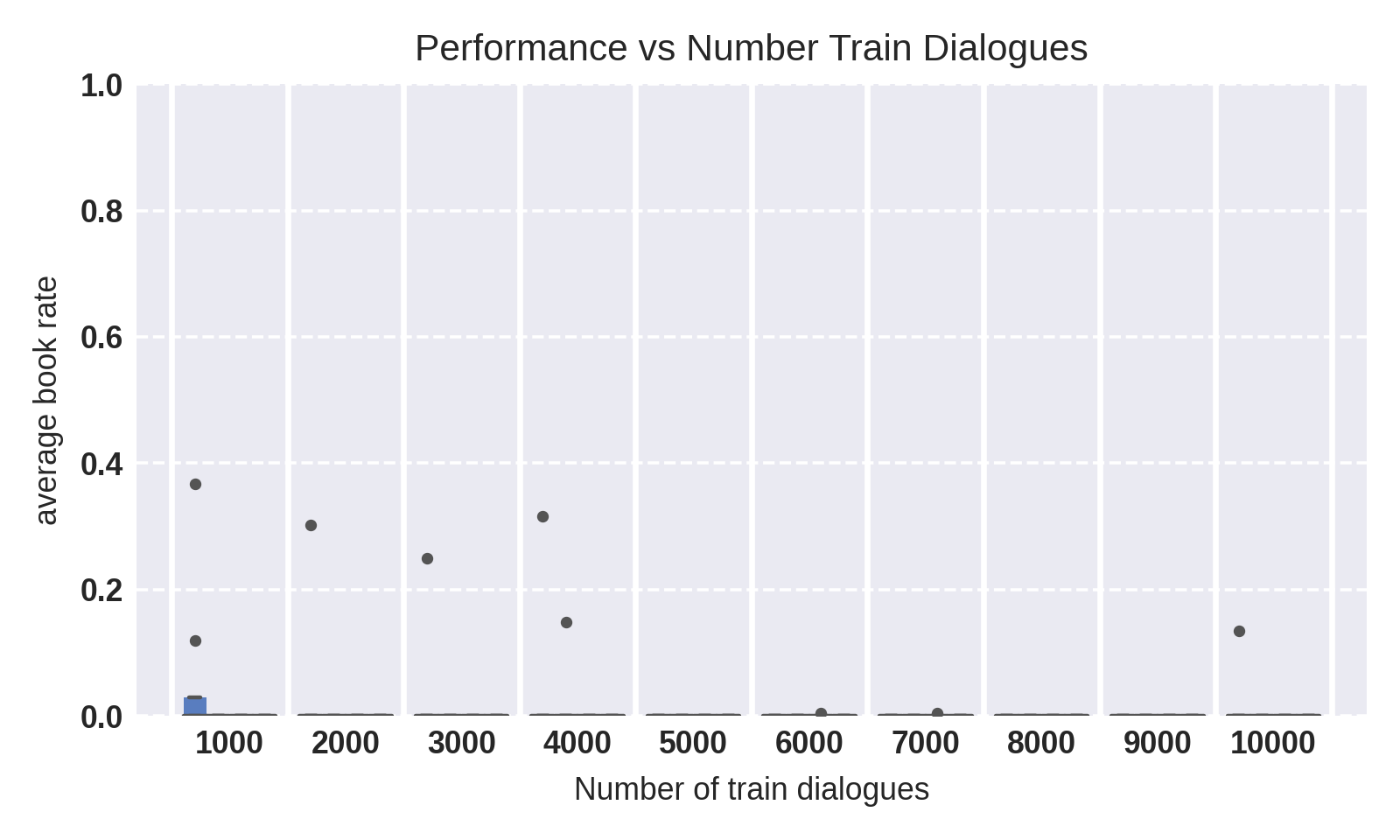}
        \label{sub:match_FNN}
    }\\
    \subfloat[Book Rate - \textsc{FNN} models with native parametrization]{
        \includegraphics[width=0.4\textwidth]{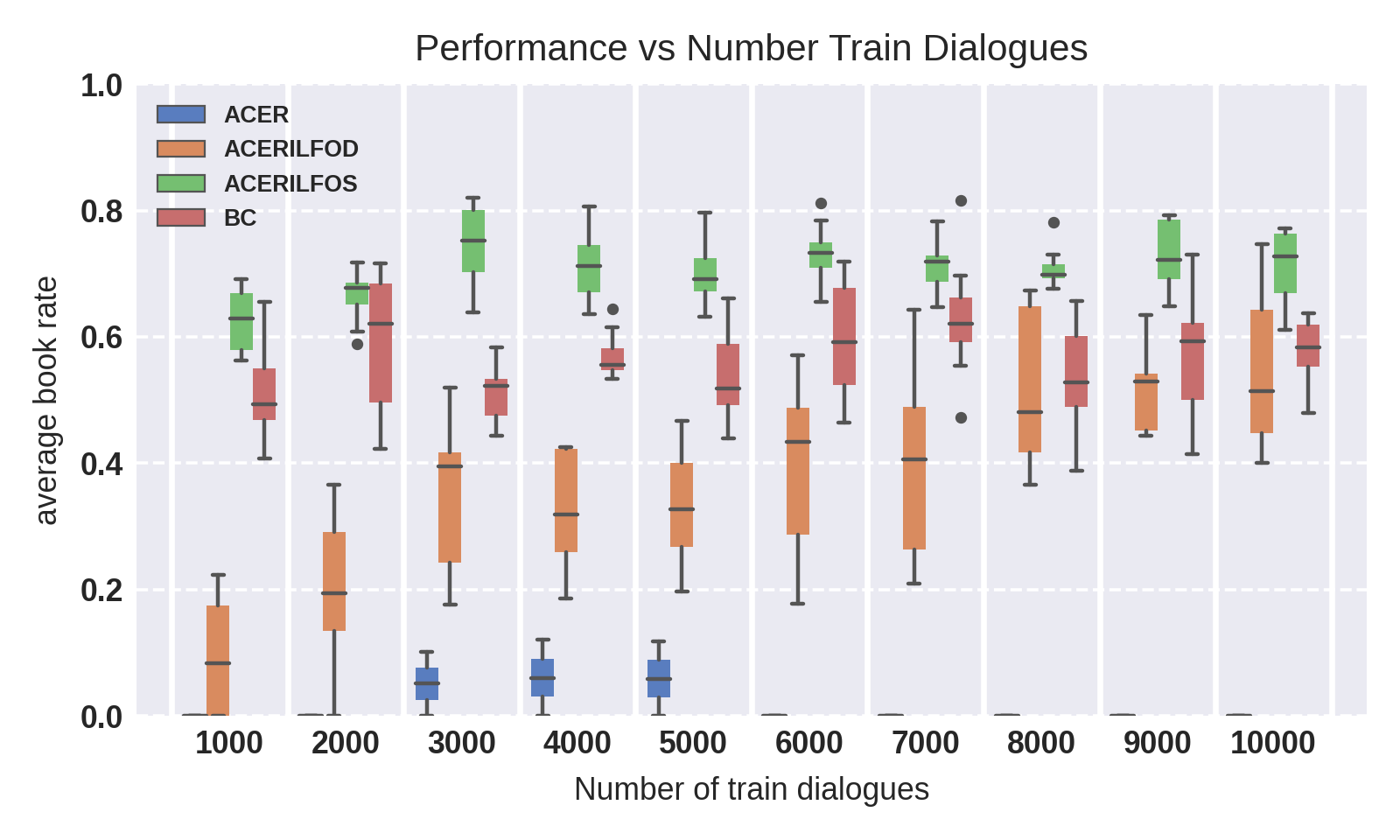}
        \label{sub:match_FNN_REF}
    }\\
  \end{center}
  \caption{Summary of performance - Task \textit{book}}
  \label{fig:all_match}
\end{figure}

\begin{figure}[ht!]
  \begin{center}
    \subfloat[Success Rate - \textsc{UHGNN} models]{
        \includegraphics[width=0.4\textwidth]{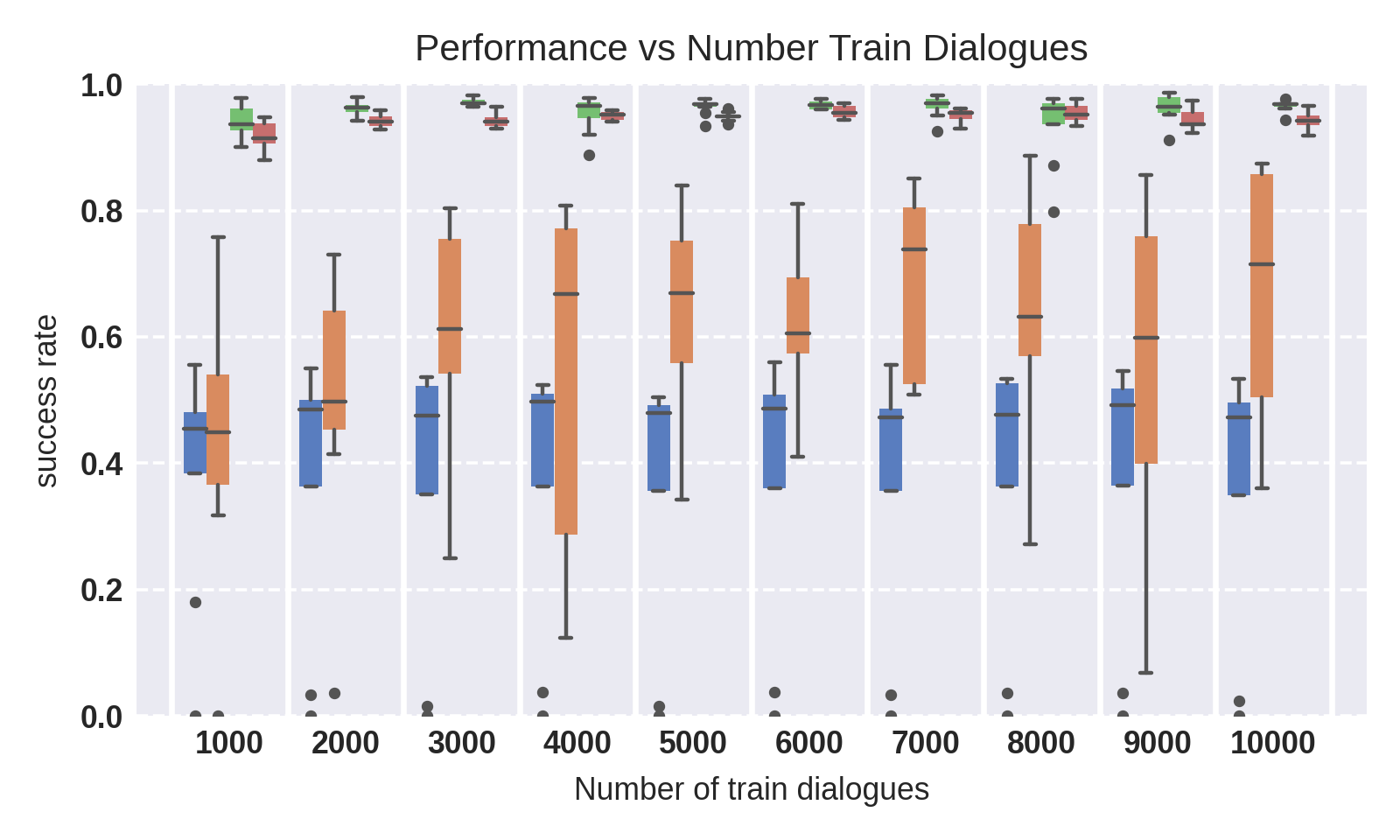}
        \label{sub:success_UHGNN}
    }\\
    \subfloat[Success Rate - \textsc{HGNN} models]{
        \includegraphics[width=0.4\textwidth]{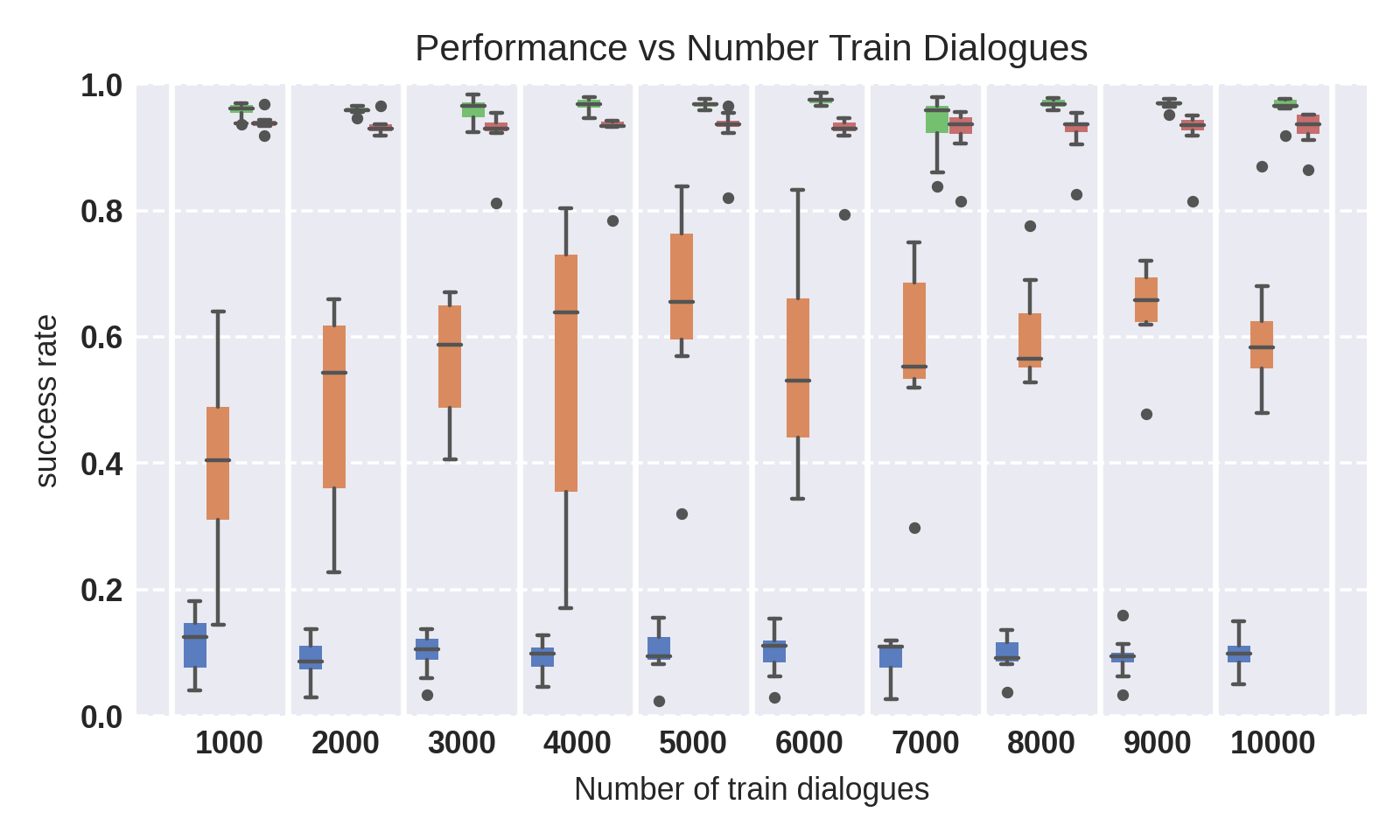}
        \label{sub:success_HGNN}
    }\\
    \subfloat[Success Rate - \textsc{HFNN} models]{
        \includegraphics[width=0.4\textwidth]{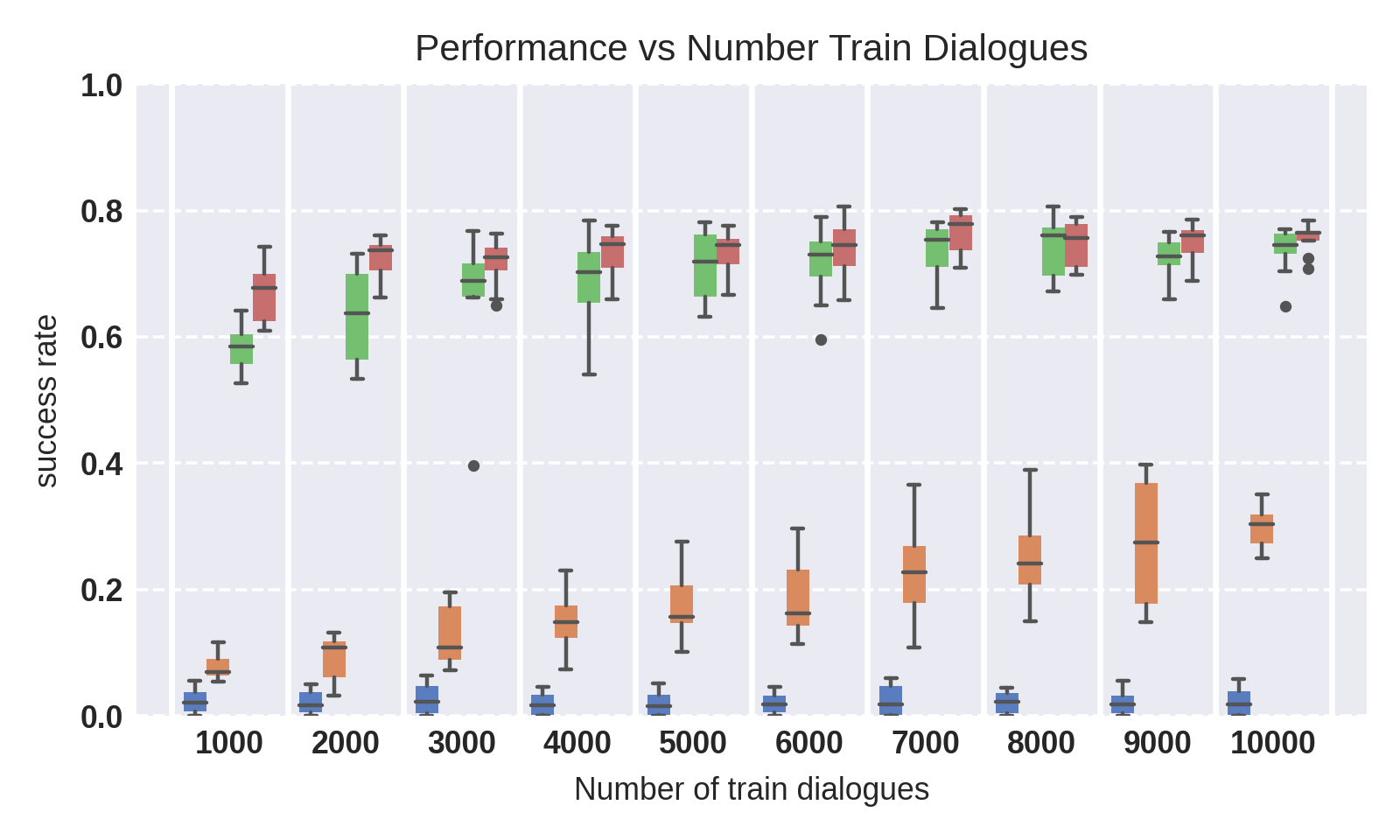}
        \label{sub:success_HFNN}
    }\\
    \subfloat[Success Rate - \textsc{FNN} models with \textsc{DIP} parametrization]{
        \includegraphics[width=0.4\textwidth]{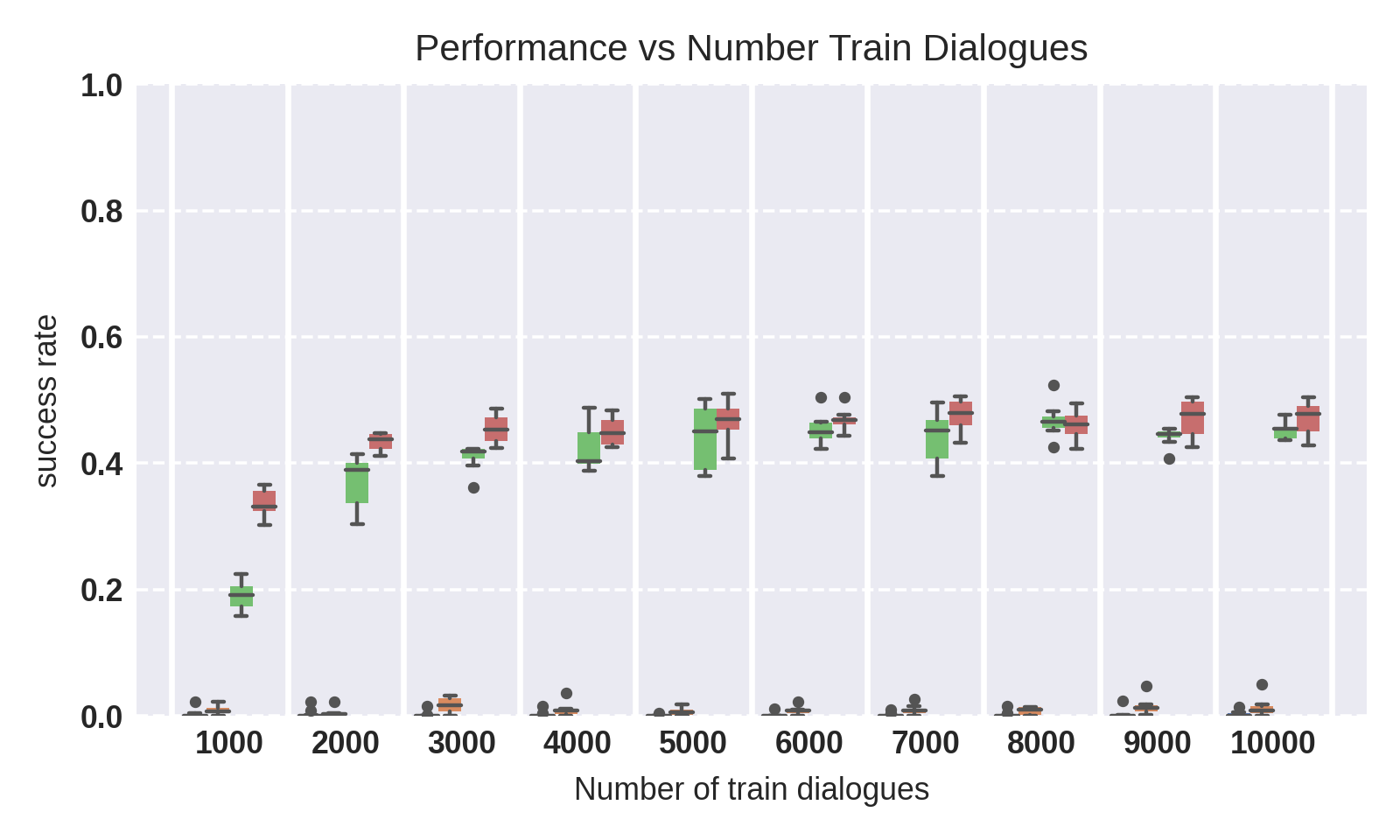}
        \label{sub:success_FNN}
    }\\
    \subfloat[Success Rate - \textsc{FNN} models with native parametrization]{
        \includegraphics[width=0.4\textwidth]{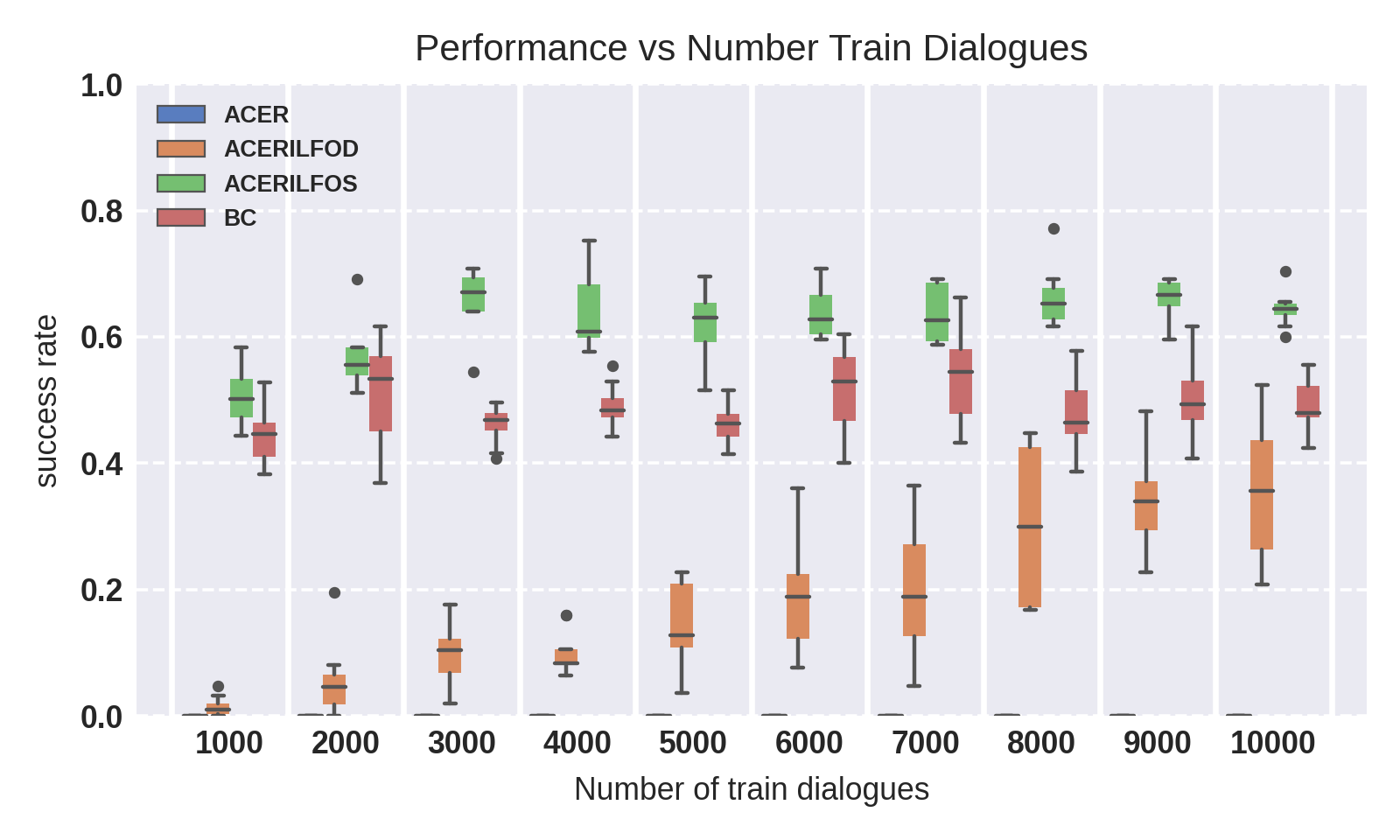}
        \label{sub:success_FNN_REF}
    }\\
  \end{center}
  \caption{Summary of performance - Global task (Task \textit{find} and/or Task \textit{book})}
  \label{fig:all_success}
\end{figure}

\begin{figure}[ht!]
  \begin{center}
    \subfloat[Cumulative rewards - \textsc{UHGNN} models]{
        \includegraphics[width=0.4\textwidth]{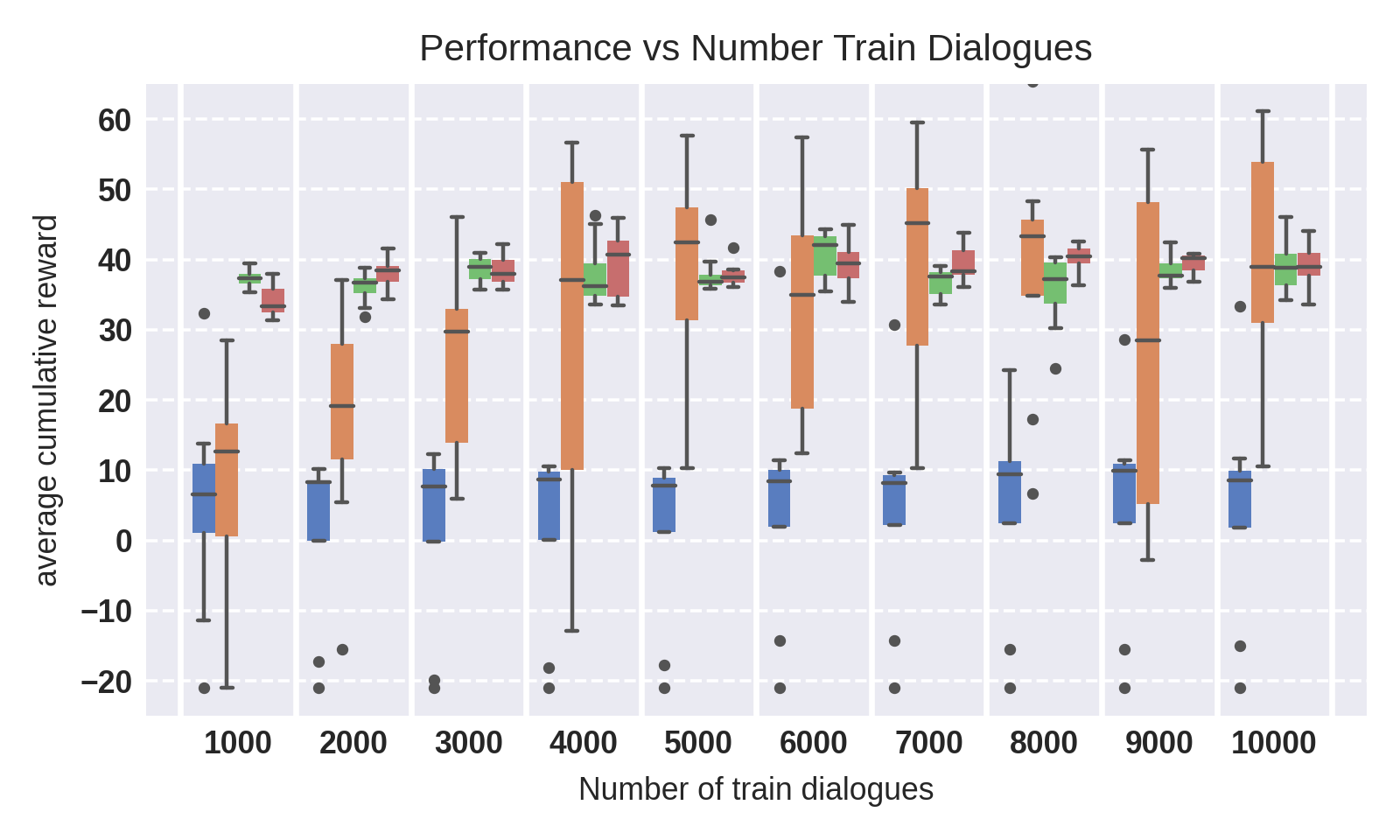}
        \label{sub:reward_UHGNN}
    }\\
    \subfloat[Cumulative rewards - \textsc{HGNN} models]{
        \includegraphics[width=0.4\textwidth]{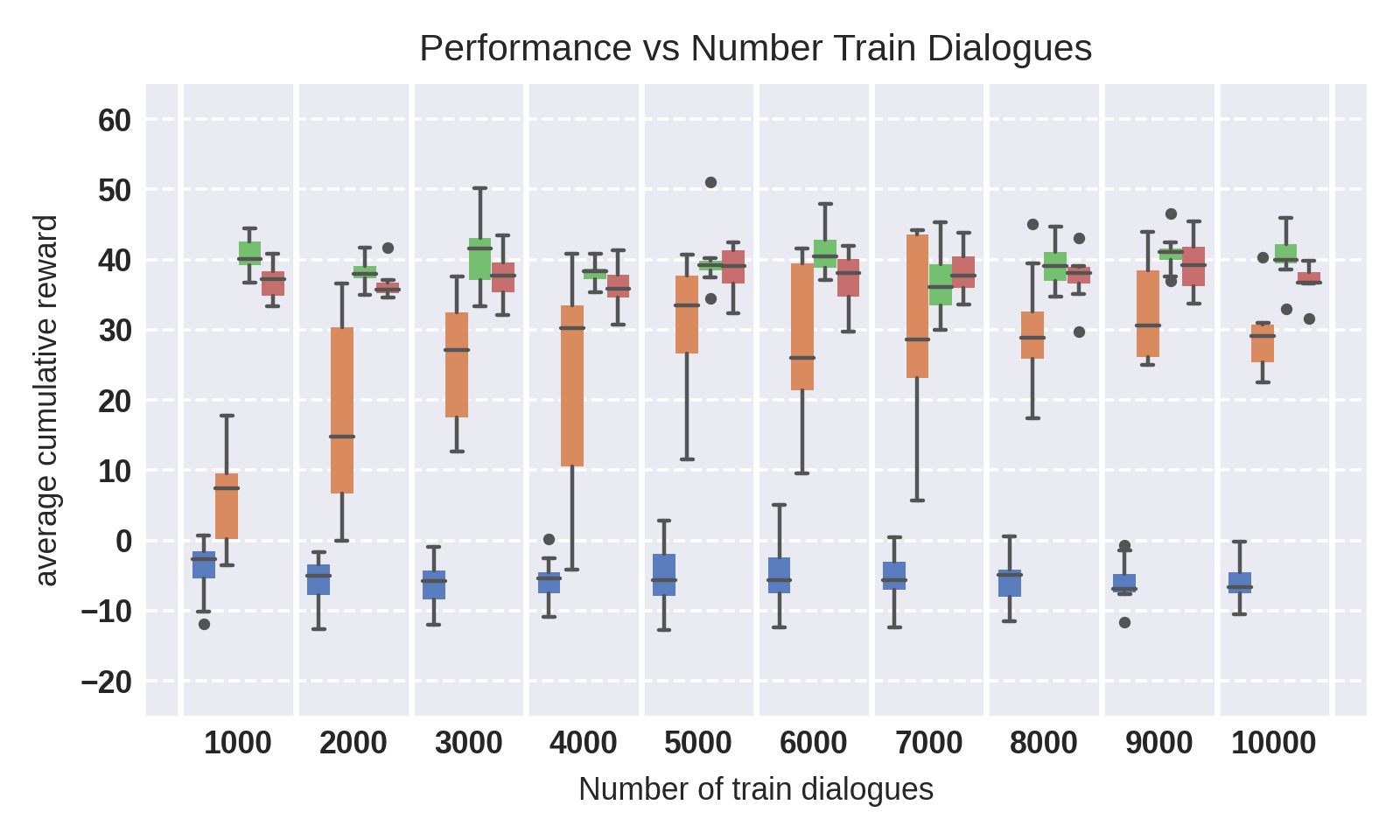}
        \label{sub:reward_HGNN}
    }\\
    \subfloat[Cumulative rewards - \textsc{HFNN} models]{
        \includegraphics[width=0.4\textwidth]{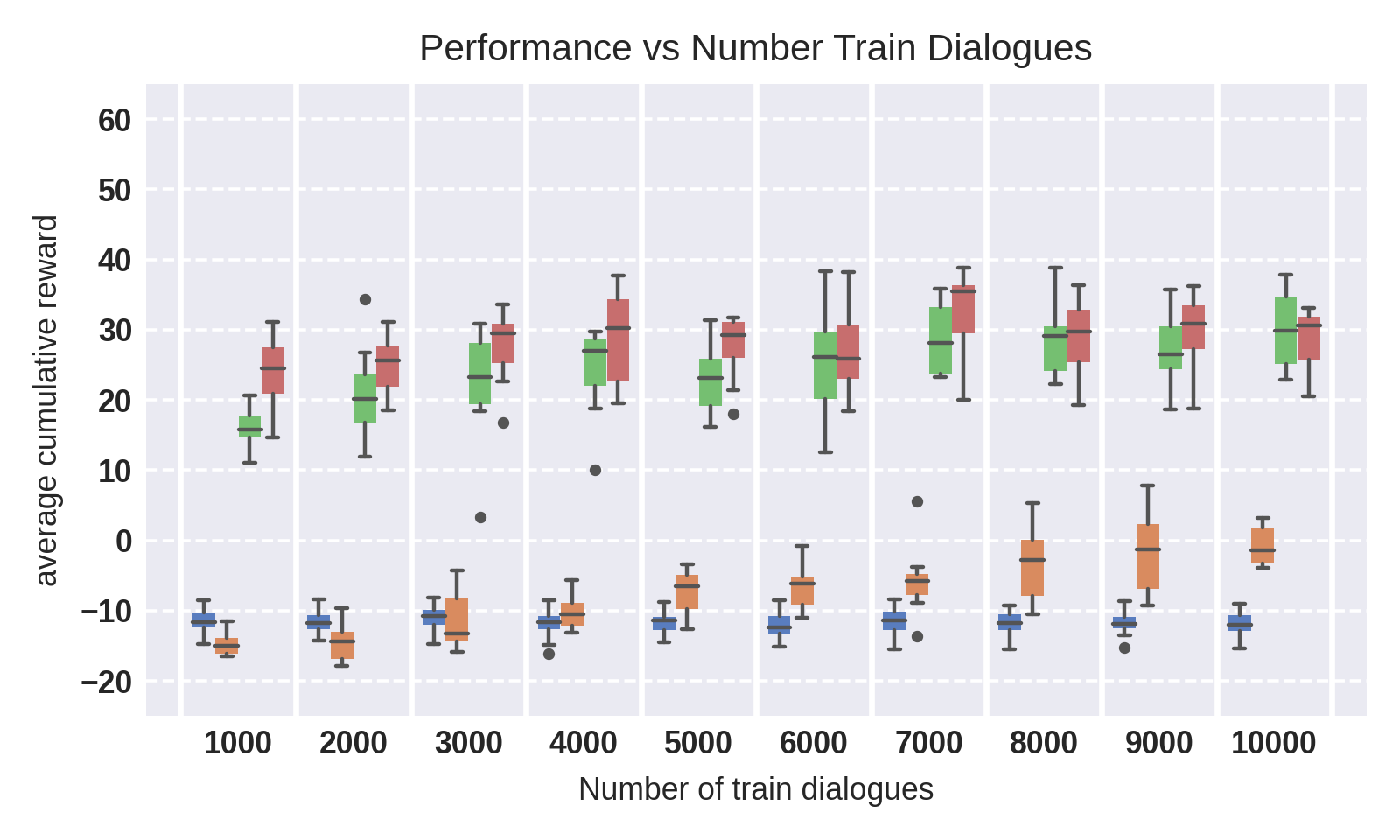}
        \label{sub:reward_HFNN}
    }\\
    \subfloat[Cumulative rewards - \textsc{FNN} models with \textsc{DIP} parametrization]{
        \includegraphics[width=0.4\textwidth]{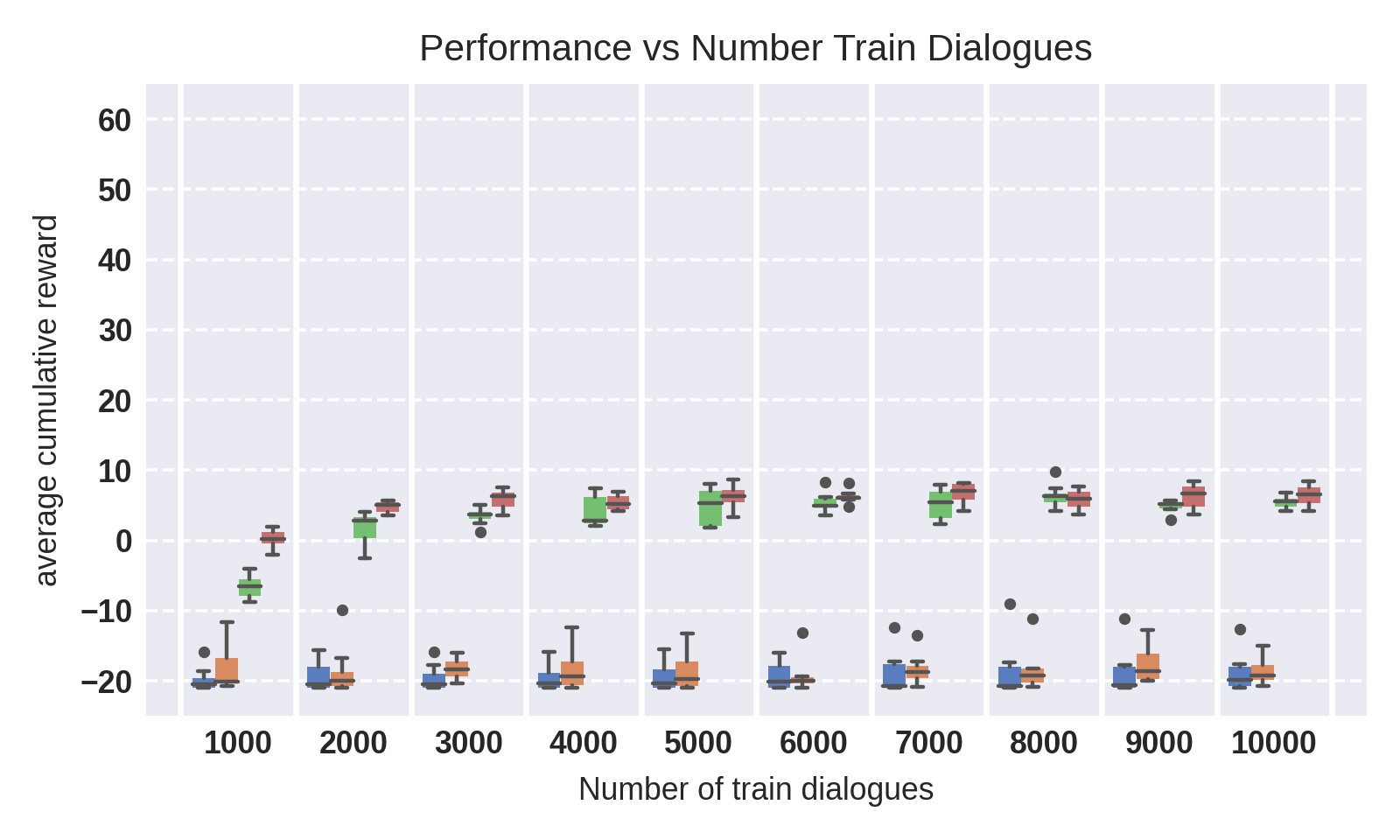}
        \label{sub:reward_FNN}
    }\\
    \subfloat[Cumulative rewards - \textsc{FNN} models with native parametrization]{
        \includegraphics[width=0.4\textwidth]{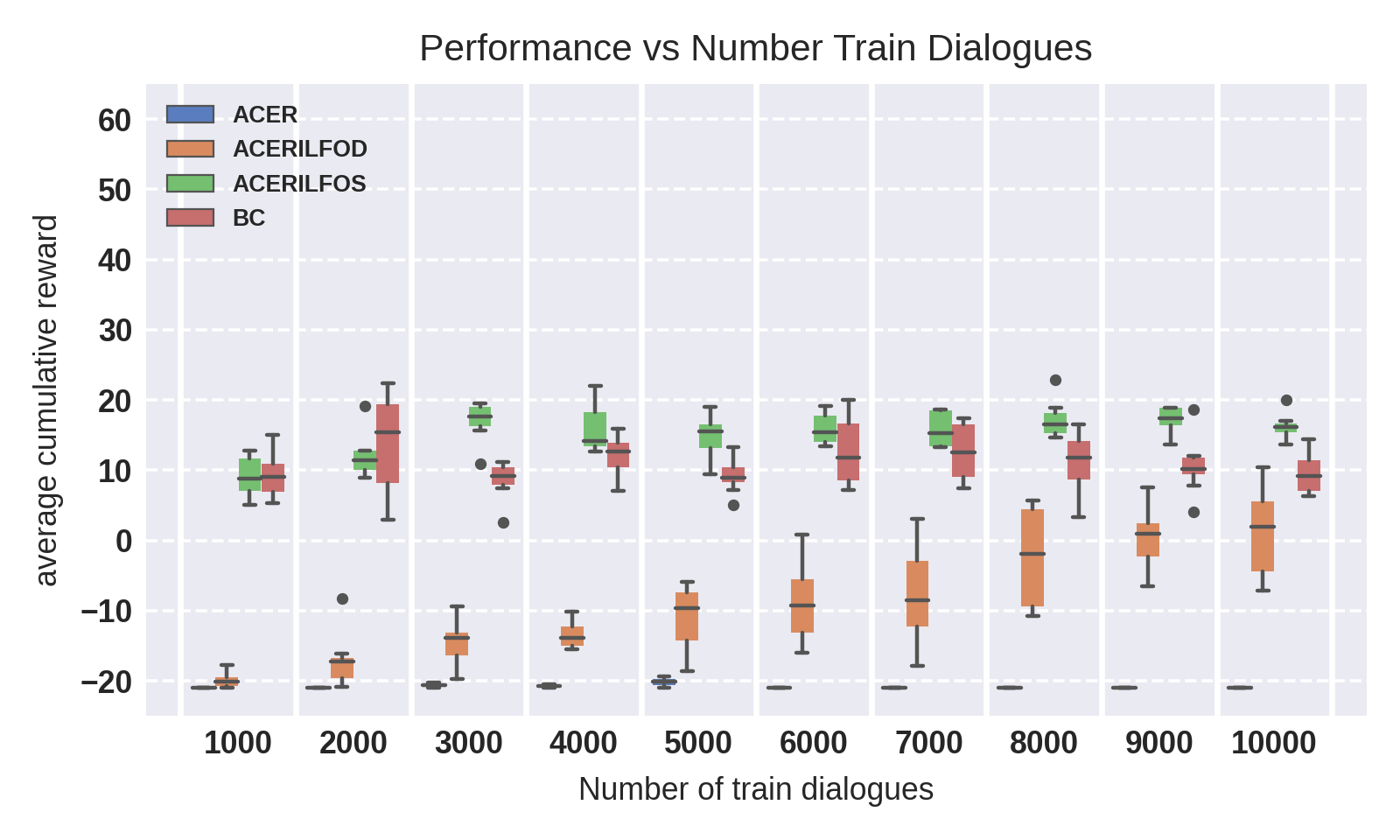}
        \label{sub:reward_FNN_REF}
    }\\
  \end{center}
  \caption{Summary of performance - Cumulative rewards}
  \label{fig:all_reward}
\end{figure}

We propose to present supplementary results of our ablation study. We show the distribution (via boxplot) of different measures with $10$ different initialisations and without pre-training. In particular, Figure~\ref{fig:all_recall} presents the distribution of inform recall, Figure~\ref{fig:all_match} the distribution of book rate, Figure~\ref{fig:all_success} the distribution of success rate and Figure~\ref{fig:all_reward} the distribution of cumulative rewards.
We precise that the coloured area represents the interquartile Q1-Q3 of the distribution, the middle line represents its median (Q2) and the points are outliers.

\end{document}